\documentclass[preprint,12pt]{elsarticle}

\usepackage{amssymb}
\usepackage{amsmath,bm}
\usepackage{graphicx}
\usepackage{booktabs}
\usepackage{multirow}
\usepackage{array,tabularx}
\usepackage[hypertexnames=false]{hyperref}
\usepackage{float}

\newcommand{\CL}{C_{L}}

\newcommand{\Ltotal}{L^{\mathrm{total}}}
\begin{document}

\begin{frontmatter}

\title{VATO: A Vortex-Force-Aware Transformer Operator for
       Unsteady Separated Aerofoil Flows}

\author[kcl]{Xingxin Yang\corref{equal}}
\author[kcl]{Zhan Zhang\corref{equal}}
\author[kcl]{Yichen Li}
\author[kcl]{Juan Li\corref{cor1}}

\cortext[equal]{These authors contributed equally to this work.}
\cortext[cor1]{Corresponding author.}
\ead{juan.li@kcl.ac.uk}

\affiliation[kcl]{organization={Department of Engineering, King's College London},
             city={London},
             postcode={WC2R 2LS},
             country={UK}}

\begin{abstract}
Accurate prediction of unsteady separated flows is challenging because the aerodynamic loads depend on nonlinear separation and vortex-shedding dynamics. Although high-fidelity CFD resolves these mechanisms, its cost limits repeated use in design and control. Standard field-level surrogate training, however, does not distinguish the flow regions that contribute most strongly to the aerodynamic loads. We introduce VATO (Vortex-Force-Aware Transformer Operator), which couples the Vortex Force Map (VFM) method to a geometry-aware neural operator through two complementary mechanisms. VATO-S adds training-only supervision of the local VFM force-contribution field, with no increase in model size or inference cost. VATO-A uses VFM contribution and sensitivity fields to prioritise force-relevant source locations for residual cross attention. The methods are evaluated on unsteady CFD data for double-edged-plate aerofoils over 54 trajectories from nine geometries. Over lead times of 1--20~ms, VATO-S reduces velocity, pressure, and vorticity errors by 10.4\%, 1.0\%, and 15.6\%, respectively, while VATO-A achieves reductions of 15.8\%, 7.5\%, and 31.2\%. VATO-S gives the lowest VFM-derived drag error, whereas VATO-A gives the lowest pressure-derived lift and drag errors. Over lead times extending 50\% beyond the training range, VATO-A retains a 26.9\% reduction in vorticity error and larger improvements in all four force readouts, despite reduced gains in velocity and pressure. These results show that force-aware operator learning can improve both flow-field prediction and aerodynamic functional accuracy in unsteady separated flows.
\end{abstract}
\begin{keyword}
neural operator \sep vortex-aware operator learning \sep
residual cross attention \sep
unsteady separated flow 
\end{keyword}

\end{frontmatter}


\section{Introduction}
\label{sec:intro}

Accurate prediction of unsteady aerodynamic loads is essential for the design,
optimisation, and control of vehicles operating in separated flow. In such
regimes, lift and drag are governed by flow separation, vortex formation, and
vortex shedding, whose strongly nonlinear and history-dependent dynamics are
difficult to represent with linearised aerodynamic models. High-fidelity
computational fluid dynamics (CFD) can resolve these mechanisms, but the cost of
repeated simulations remains prohibitive for iterative design and control. This
has motivated the development of data-driven surrogate models that approximate
the evolution of the flow and its aerodynamic response at substantially reduced
computational cost.

Machine learning has been widely applied across fluid mechanics over the past
decade~\cite{ref01,ref02}, while data-driven modelling of unsteady aerodynamics
has developed into a substantial research area of its own~\cite{ref03}. 
Existing approaches can broadly be separated according to whether they predict
aerodynamic quantities directly or reconstruct the underlying flow field. At the level of aerodynamic coefficients, Yao
et al.~\cite{ref04} optimised the endurance coefficient of a tail-sitter UAV
using a multi-objective genetic algorithm, Lou et al.~\cite{ref05} modified
aerofoil geometry with a double deep Q-network using a neural-network reward,
Zhu et al.~\cite{ref06} learned a closure between turbulent eddy viscosity and
mean flow variables, and Zhao et al.~\cite{ref07} predicted the lift-to-drag
characteristics of Mars helicopter aerofoils. 
Liu et al.~\cite{ref08} review this broader class of surrogate-based
aerodynamic shape modelling and optimisation. These approaches demonstrate that
learned surrogates can replace repeated solver evaluations when integral
aerodynamic quantities are the primary outputs, but models based on steady or
time-averaged quantities do not directly resolve the unsteady flow structures
responsible for transient load histories.
Field-level surrogates instead seek to reconstruct the spatial flow state from
which aerodynamic quantities can subsequently be evaluated.
Convolutional and
fully connected architectures have reproduced aerodynamic fields and
coefficients across varied geometries~\cite{ref09,ref10,ref11}, while recurrent
architectures have been developed for unsteady modelling~\cite{ref12,ref13,ref14}
and high-incidence prediction~\cite{ref15}.
Recursive temporal prediction,
however, remains susceptible to accumulated error over long horizons.
Physics-informed formulations constrain the learned field by the governing
equations~\cite{ref16}. Neural operators learn mappings between function spaces
that transfer across discretisations, through branch-trunk architectures built
on the universal approximation theorem for operators~\cite{ref17}, spectral
parameterisations of the integral kernel~\cite{ref18}, and formulations that
impose boundary conditions for aerofoil flows~\cite{ref19}. Physics-guided
reconstruction has recovered high-fidelity fields from sparse or low-resolution
inputs with diffusion models~\cite{ref20,ref21} and regularised sparse
compressible-flow reconstruction with discrete conservation
residuals~\cite{ref22}.Geometry-aware operator transformers further
extend operator learning to unstructured meshes and non-uniform domains~\cite{ref23},
enabling a single surrogate to operate across families of geometries.

Despite these advances, the training signal in field-level surrogates is applied
uniformly over the computational domain. A relative error on velocity, pressure,
or vorticity assigns every mesh point the same weight, so the wake and far
field, which occupy most of the mesh, dominate the objective, while the
aerodynamic load originates in a small fraction of the domain. The objective
therefore contains no mechanism that identifies the flow structures on which the
downstream design or control task depends.
The Vortex Force Map (VFM) method~\cite{ref24,ref25,ref26} provides a
mechanics-based means of making this distinction. VFM solves an auxiliary
potential problem determined by the body geometry and force direction, producing
a vortex-force factor field whose interaction with the local velocity and
vorticity identifies the spatial contribution of the flow to lift or drag. 
The
auxiliary field requires no additional flow solution and can therefore be
precomputed for each geometry and incidence before operator training. 
VFM was introduced to estimate unsteady forces from measured or computed
snapshots~\cite{ref24,ref25} and has subsequently been extended to three-dimensional (3-D) unsteady
flow~\cite{ref26}. He et al.~\cite{refFCN} combined
vortex-force information with a graph convolution attention network to infer
velocity and vortex-force contributions from incomplete flow measurements and
thereby recover force coefficients for flow around a circular cylinder, demonstrating that vortex-force information can guide reconstruction from
sparse observations. However, the question remains: whether vortex-force awareness can be introduced to shape the learning and information-routing
mechanisms of a neural operator for full-field unsteady prediction. 

To address this question, we introduce VATO (Vortex-Force-Aware Transformer Operator), a family of neural operators
that couples the VFM method to a geometry-aware transformer
backbone~\cite{ref23} at two interfaces. VATO-S supervises the per-point VFM
contribution field during training, leaving the architecture, the parameter
count, and the inference-time operator of the backbone unchanged, and therefore
suits deployments in which inference cost is fixed. VATO-A uses the current flow
vorticity together with the VFM contribution and sensitivity fields to identify
regions relevant to Lift and Drag, and admits the ordinary flow state from those
regions through residual cross attention without embedding the VFM values,
providing larger field improvements where additional inference cost is
acceptable. Both configurations are trained with a flow-aware sampler, and a
sampling-matched control is retained alongside the retrained backbone reference
so that the effect of each coupling is separated from the effect of the sampler.
Neither coupling is tied to a particular section: the vortex-force fields follow
from the shape and the incidence alone, and the backbone accepts arbitrary
unstructured meshes.

The double-edged-plate (DEP) aerofoil proposed for Martian rotorcraft provides a demanding test case for this purpose. The thin Martian atmosphere places rotor blades at low
Reynolds numbers of order $10^{4}$-$10^{5}$, where leading-edge separation bubbles form
readily~\cite{ref27}, may burst into fully separated
states~\cite{ref28}, and the separation location in this regime materially
affects drag~\cite{ref29}. Sharp-edged sections address this sensitivity by
imposing the separation point geometrically~\cite{ref30}, so that the load is
governed by the vortex structures rather than by the onset of separation. Traub
and Coffman~\cite{ref31} showed experimentally that leading- and trailing-edge
folds fix the location at which separation begins and improve the efficiency of
thin plate aerofoils at low Reynolds number, and Koning et al.~\cite{ref32}
evaluated and optimised the resulting DEP family for Martian rotor
applications. 
Varying the two fold angles produces a family of geometries with the same separation mechanism but different vortex structures and aerodynamic responses. This provides a suitable test case for force-aware operator learning.

The main contributions of this work are as follows.
\begin{enumerate}
\item A training-only coupling of the VFM method to a neural operator, VATO-S,
      which supervises where the aerodynamic load originates and improves field
      prediction at the parameter count and inference cost of the backbone.
\item An architectural coupling, VATO-A, in which VFM contribution and
      sensitivity fields identify Lift- and Drag-relevant regions whose flow
      state enters residual cross attention, yielding the strongest field
      accuracy of the family, most prominently in vorticity.
\item An attribution protocol built on a retrained reference and a
      sampling-matched control, which separates the effect of the vortex-force
      couplings from that of the training sampler and shows the two act on
      different quantities.
\item An evaluation across held-out incidences and a 50\% lead-time extension,
      in which ($C_L$) and ($C_D$) are recovered from the predicted fields through two
      independent readouts, a pressure-surface integral and the VFM volume
      integral.
\end{enumerate}

The remainder of this paper is organised as follows. Section~\ref{sec:dataset}
describes the DEP dataset and its CFD configuration. Section~\ref{sec:method}
presents the backbone, the two couplings, and the training protocol.
Section~\ref{sec:results} reports the benchmark, the force functionals, and the
predicted fields. Section~\ref{sec:discussion} discusses the findings and their
limitations, and Section~\ref{sec:conclusion} concludes.

\section{Dataset}
\label{sec:dataset}
The dataset consists of two-dimensional unsteady CFD solutions for a family of
DEP aerofoils, spanning fourteen geometries and thirteen angles of attack.
\subsection{Geometry and CFD configuration}
\label{subsec:geometry}

\begin{figure}[H]
  \centering
  \includegraphics[width=\linewidth]{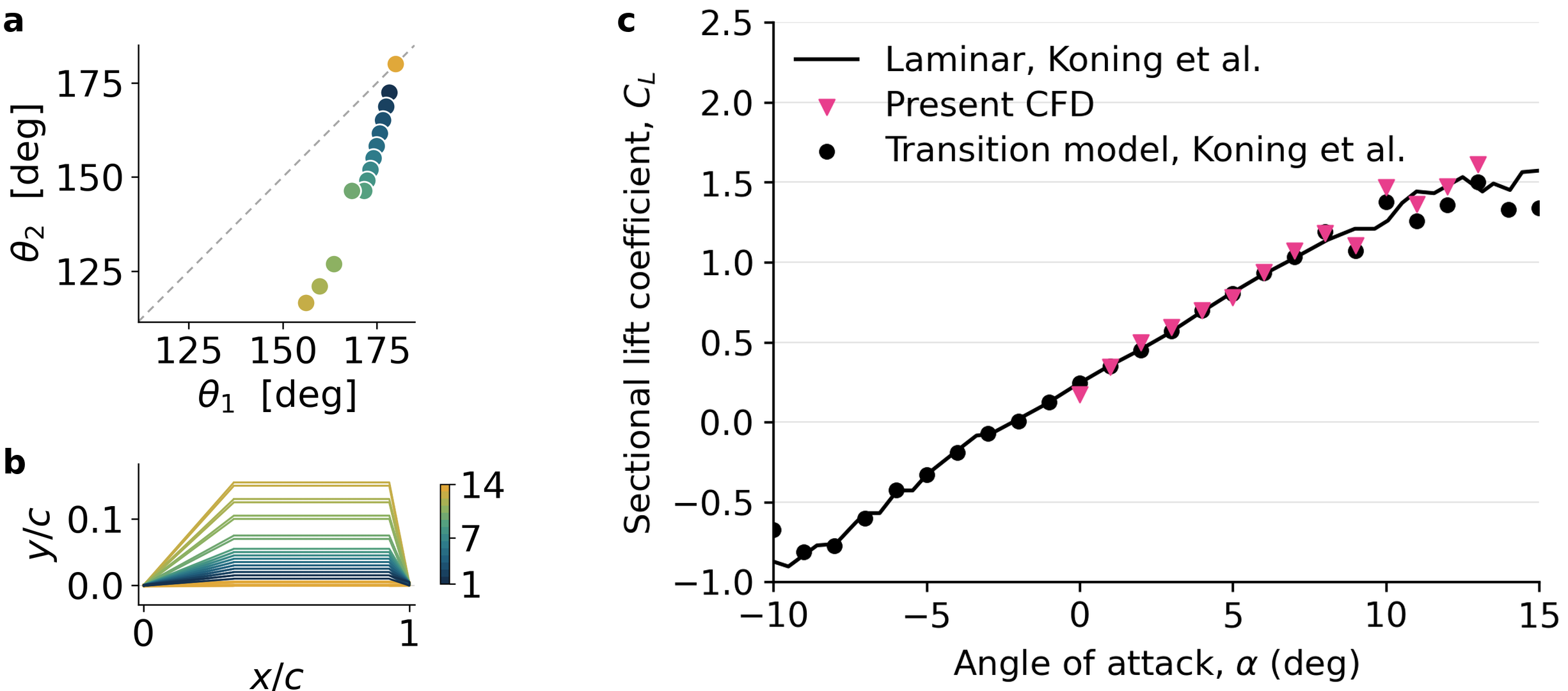}
  \caption{Dataset geometry and CFD cross-check. (a)~Envelope in
$(\theta_{1},\theta_{2})$ interior-fold-angle space; the dashed diagonal denotes
$\theta_{1}=\theta_{2}$. (b)~Chord-normalised surface contours of the fourteen
DEP geometries, coloured by geometry index. (c)~Time-averaged sectional $\CL$ from the present STAR-CCM+ simulations, compared with the
transition-model and fully laminar results computed by Koning et
al.~\cite{ref32} on the matched configuration.}
  \label{fig:dataset_geometry_validation}
\end{figure}

The DEP aerofoil is parameterised by independent leading- and trailing-edge fold
angles, $\theta_1$ and $\theta_2$, measured as the interior angle at each fold
vertex, so that $180^\circ$ denotes an unfolded edge and a smaller angle denotes
a larger deflection of the surface at that vertex. Fourteen geometries span a
continuous envelope from the most folded shape ($\theta_1 = 156.2^\circ$,
$\theta_2 = 116.6^\circ$) to the flat-plate reference ($\theta_1 = \theta_2 =
180^\circ$), shown in Figure~\ref{fig:dataset_geometry_validation}a,b.

Each geometry and angle-of-attack pair was solved as a two-dimensional unsteady
case in STAR-CCM+ on an unstructured mesh of 39{,}552 to 44{,}600 cells,
corresponding to 24{,}404 to 29{,}178 mesh points. The Martian atmospheric
conditions are a density of $0.01459\,\mathrm{kg\,m^{-3}}$ and a dynamic
viscosity of $1.22\times10^{-5}\,\mathrm{Pa\,s}$, and the freestream speed of
$69.6\,\mathrm{m\,s^{-1}}$ follows Koning et al.~\cite{ref32}, giving a chord
Reynolds number of 10{,}117. Turbulence is treated with the SST $k$--$\omega$
model. The computational domain extends five chords upstream and eight chords
downstream of the section, and time integration uses a physical time step of
$5\times10^{-5}\,\mathrm{s}$.

Sampling began at initialisation and continued until the $C_L$ and $C_D$ settled into a regular periodic oscillation, which required between
4,000 and 10,000 time steps depending on the geometry and the angle of attack.
Recording every 20 steps gives trajectories of 200 to 500 frames at intervals of
1~ms, each containing the initial transient
followed by the established shedding. Velocity and pressure were exported at every sampled step. The angle of attack ranges from $1^\circ$ to
$13^\circ$.

The CFD setup was cross-checked against the independent Mars-rotorcraft baseline
of Koning et al.~\cite{ref32} on a matched flat-plate and DEP configuration.
Their computations solve the compressible RANS equations in OVERFLOW~2.2o with
the SA-neg-1a one-equation model coupled to the AFT2017b transition model, and a
second set of cases is run fully laminar. The two baselines agree closely with
each other over the operating envelope of interest, so at this Reynolds number
the sectional lift is set by the geometrically fixed separation rather than by
the turbulence closure. The present time-averaged $C_L$ agree with
both baselines across that envelope
(Figure~\ref{fig:dataset_geometry_validation}c).

\subsection{Dataset training settings}
\label{subsec:split}

The dataset spans fourteen geometries at thirteen angles of attack: nine
geometries were simulated across the full incidence range and five at
$1^\circ$, $6^\circ$, and $12^\circ$ only, giving 132 trajectories. Seven
incidences provide 68 training trajectories, on which the lead time is sampled
at random from 1 to 20~ms, and four further incidences provide 36 validation
trajectories evaluated at a fixed lead time of 1~ms to monitor the field
objective during training; all reported comparisons use the final epoch-300
checkpoint, so no checkpoint was selected on the validation set. The test set
contains 54 trajectories from the nine fully simulated geometries at six
incidences, none of which appears in training, and of these $6^\circ$ and
$12^\circ$ appear in neither the training nor the validation set and therefore
test interpolation to an unseen angle of attack. Evaluation covers every integer
lead time from 1 to 30~ms, extending 50\% beyond the longest lead time covered
in training and testing extrapolation in lead time (Table~\ref{tab:split}).

\begin{table}[htbp]
  \centering
  \caption{Data split by angle of attack. Lead times are sampled at random within
the stated range during training and evaluated at every integer value within it
at test time. The validation set serves only to monitor the field objective and
is evaluated at a single lead time.}
  \label{tab:split}
  \setlength{\tabcolsep}{9pt}
  \large
  \resizebox{\textwidth}{!}{%
  \begin{tabular}{llcc}
    \toprule
    Split      & Angles of attack & Lead time $\tau$ (ms)
               & Trajectories \\
    \midrule
    Train      & $1^{\circ},2^{\circ},4^{\circ},7^{\circ},
                  8^{\circ},10^{\circ},13^{\circ}$ & 1--20 & 68 \\
    Validation & $3^{\circ},5^{\circ},9^{\circ},11^{\circ}$ & 1 & 36 \\
    Test & $3^{\circ},5^{\circ},6^{\circ},9^{\circ},
            11^{\circ},12^{\circ}$ & 1--30 & 54 \\
    \bottomrule
  \end{tabular}
  }
\end{table}

\section{Method}
\label{sec:method}

\subsection{GAOT backbone and VATO variants}
\label{subsec:backbone}

All four reported configurations use the GAOT backbone~\cite{ref23}. The backbone consists of three core modules: a multiscale attentional graph neural operator (MAGNO) encoder, which maps unstructured physical coordinates onto a regular latent grid; a vision transformer, which divides the latent grid into patch tokens and processes them with global attention; and a MAGNO decoder, which maps the processed latent features back onto arbitrary query coordinates. In the validated configuration, the latent grid size is $128 \times 128$, the lifting channel dimension is 96, the number of transformer layers is six, the number of 
attention heads is eight, the MAGNO neighbourhood radius is 0.03 with multiscale
factors $[0.5, 1.0]$, and at most 512 neighbours are aggregated per query point.

\begin{figure}[H]
  \centering
  \includegraphics[width=\linewidth]{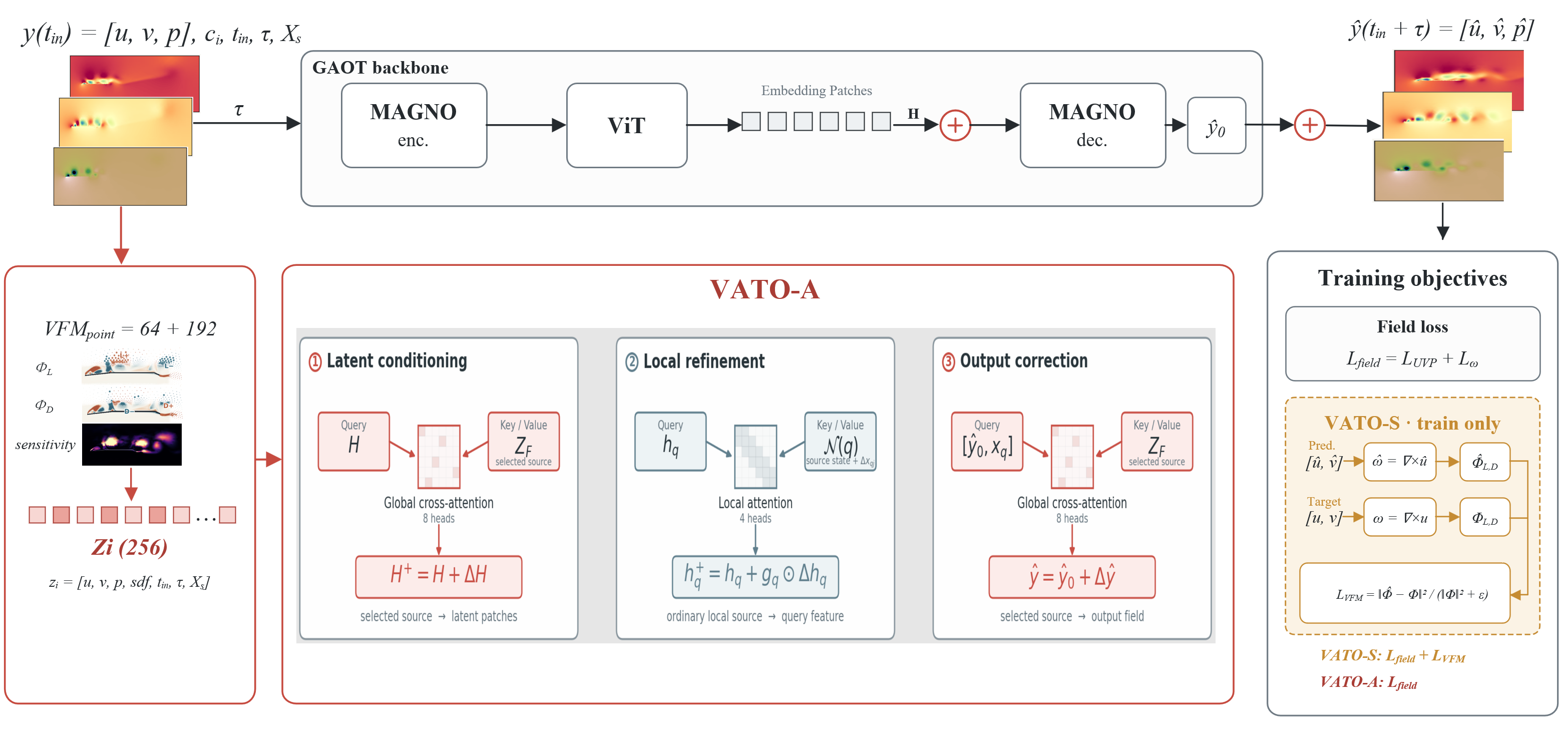}
  \caption{VFM coupling to the shared GAOT backbone. A parameter-free
prioritisation rule forms 256 ordinary source tokens from the current input
frame: 64 indices retain geometric coverage, while the remaining 192 are
balanced over signed Lift and Drag contribution groups and VFM sensitivity. VFM
values determine the selected locations but are not embedded in the token
content. VATO-A augments the backbone through three zero-initialised residual
paths. Processed latent patches read the selected tokens through an eight-head,
width-192 cross attention; each output query reads up to 32 nearby ordinary
source points through a gated four-head, width-96 local attention with radius
0.04; and each preliminary output query reads the selected tokens through a
second eight-head, width-192 cross attention. The common field objective
combines physical-space UVP and mesh-curl signed-log-vorticity losses. VATO-S
retains the GAOT architecture and additionally applies the training-only VFM
contribution field loss, whereas VATO-A uses the field objective without that
VFM loss.}
  \label{fig:model}
\end{figure}

Two common adaptations are used for the present dataset. First, the encoder and decoder operate on different point sets, so that the model accepts a sampled source state and predicts on the native CFD mesh. Second, geometry and time are provided through a signed-distance field and temporal scalars, as defined in Section~\ref{subsec:task}. The vortex-force information is then coupled to this backbone along two routes, which define the two variants (Figure~\ref{fig:model}). VATO-S retains the architecture above and changes only the training objective, through the contribution field supervision defined in Section 3.6. VATO-A instead inserts three zero-initialised residual attention paths. After transformer processing, each latent patch reads 256 ordinary source tokens selected from the full current frame using the VFM contribution and sensitivity fields. Within the MAGNO decoder, each output query reads nearby ordinary source points through gated local attention. After the preliminary prediction, each native-mesh query reads the same 256 selected tokens again through a global cross attention residual. The VFM values determine which full-frame locations are prioritised but are not embedded in the selected token content.

\subsection{Prediction task}
\label{subsec:task}

The prediction target is the future full-mesh field
$\mathbf{y}(t_{\mathrm{in}}+\tau,\mathbf{X})$. Each model writes the output
as
\begin{equation}
  \hat{\mathbf{y}}(t_{\mathrm{in}}+\tau,\,\mathbf{X})
  = G_{\theta}\!\left(\mathbf{y}(t_{\mathrm{in}},\mathbf{X}_s),\,
    \mathbf{c}(\mathbf{X}_s),\,t_{\mathrm{in}},\,\tau,\,\mathbf{X}\right),
  \label{eq:task}
\end{equation}
where $\mathbf{X}=\{\mathbf{x}_i\}_{i=1}^{N}$ is the native CFD mesh,
$\mathbf{X}_s\subset\mathbf{X}$ is the 12{,}000-point input set used during training,
$t_{\mathrm{in}}$ is the source time, and
$\tau=t_{\mathrm{target}}-t_{\mathrm{in}}$ is the requested scalar
lead time. At each mesh point, the state vector is
\begin{equation}
  \mathbf{y}_i(t)
  =
  [u_i(t),v_i(t),p_i(t)]^{T}.
  \label{eq:state_field}
\end{equation}
Here $u_i$ and $v_i$ are velocity components in m\,s$^{-1}$ and $p_i$ is
pressure in Pa. The spatial condition field $\mathbf{c}$ contains the signed
distance to the aerofoil surface,

The input at each source point is $[u,v,p]^{T}$ together with $c$ and
the temporal scalars $t_{\mathrm{in}}$ and $\tau$ (Figure~\ref{fig:model}).

Because the CFD mesh is strongly refined near the body, with roughly 72\% of
points lying within a near-wall band of signed distance $\le 0.1c$, an
unstratified draw of encoder points would under-represent the wake. The sampler
therefore reserves 6\% of the encoder budget for body-wall points, caps the
near-wall band at 50\%, and assigns the remaining 44\% to the far field, while
in the reported benchmark every configuration uses $\mathbf{X}_s = \mathbf{X}$
as its encoder source. Training uses direct single-step prediction, mapping a
source state to a single requested lead time rather than through a recursive
rollout.

\subsection{Pressure surface integral}
\label{subsec:pressure}

$C_L$ and $C_D$ are recovered from a predicted field through two readouts, The first integrates the
pressure over the body wall. Let $\Omega_b$ denote the solid aerofoil region and $\partial\Omega_b$ its
boundary. With $\mathbf{n}$ the unit normal directed from the body into the
fluid, the pressure force per unit span is
\begin{equation}
  \mathbf{F}^{\mathrm{press}}
  =
  -\int_{\partial\Omega_b}
  p\,\mathbf{n}\,ds .
  \label{eq:F_press}
\end{equation}
The Cartesian components are projected onto the wind-axis drag and lift
directions using the angle of attack $\alpha$:
\begin{align}
  F_D^{\mathrm{press}}
  &=
  F_x^{\mathrm{press}}\cos\alpha
  +
  F_y^{\mathrm{press}}\sin\alpha,
  \label{eq:F_press_drag}\\
  F_L^{\mathrm{press}}
  &=
  -F_x^{\mathrm{press}}\sin\alpha
  +
  F_y^{\mathrm{press}}\cos\alpha .
  \label{eq:F_press_lift}
\end{align}
With freestream density $\rho_\infty$, speed $U_\infty$, and reference chord
$c$, the corresponding pressure-induced force coefficient is
\begin{equation}
  C_k^{\mathrm{press}}
  =
  \frac{F_k^{\mathrm{press}}}
  {\frac{1}{2}\rho_\infty U_\infty^2 c},
  \qquad k\in\{L,D\}.
  \label{eq:Ck_press}
\end{equation}
In the implementation, Eq.~\eqref{eq:F_press} is evaluated by quadrature over
the body-boundary edges of each sample's native mesh. For the pressure
readout, $p$ is given by either the target or model-predicted pressure field.

\subsection{VFM volume integral}
\label{subsec:vfm}

\begin{figure}[H]
  \centering
  \includegraphics[width=\linewidth]{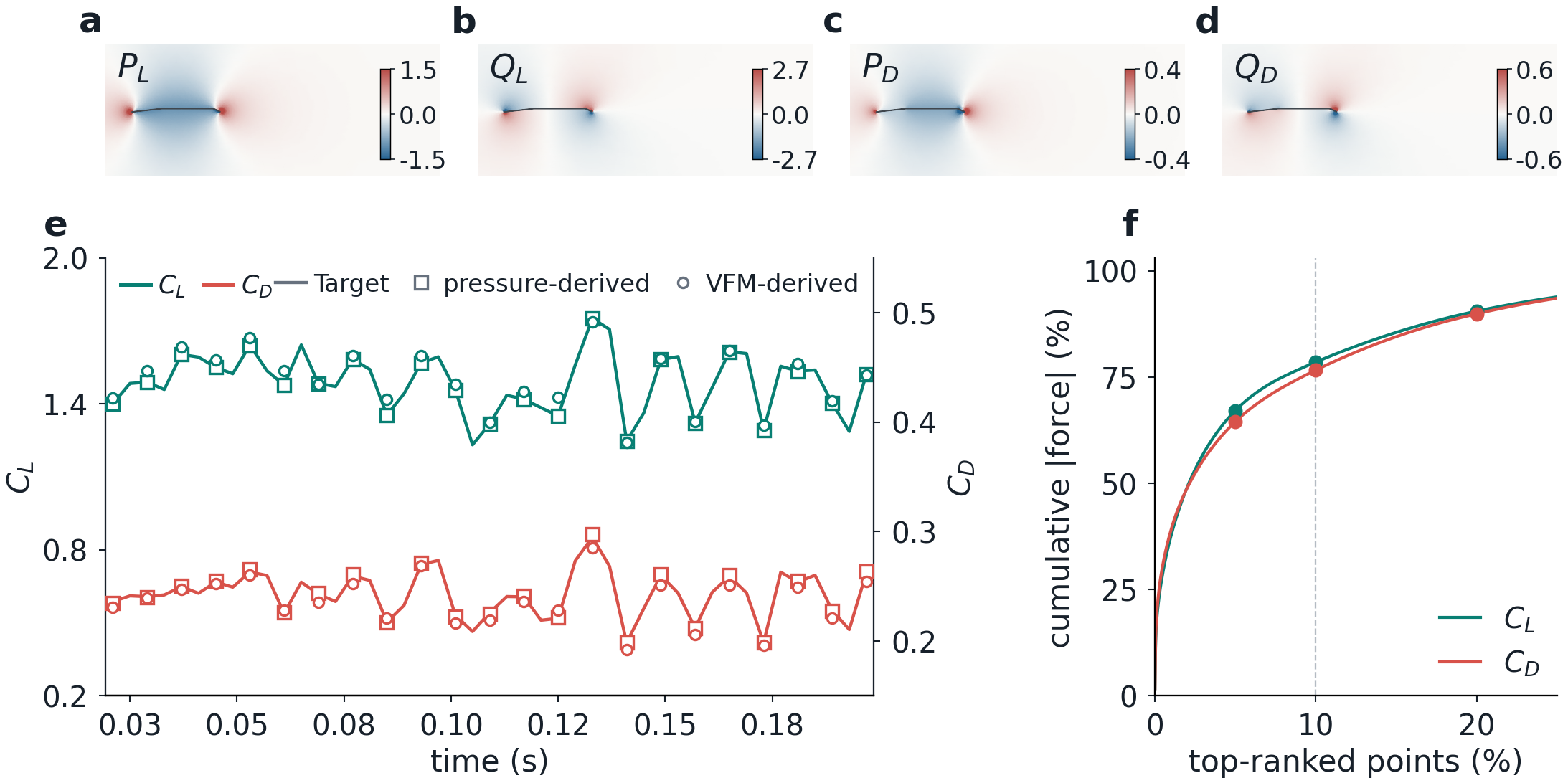}
  \caption{Physical basis for the VFM coupling, shown for
    $(\theta_1,\theta_2)=(174.1^{\circ},155.0^{\circ})$ at
    AoA $=12^{\circ}$. (a--d)~Panel-method coefficient fields
    $P_L,Q_L,P_D,Q_D$. (e)~Lift and drag recovered on paired left/right ordinates from the target force record,
    the pressure integral on the target pressure field, and the VFM integral
    evaluated on the target velocity and mesh curl; the VFM mean absolute relative errors
    are 2.09\% for $C_L$ and 2.22\% for $C_D$. (f)~Cumulative absolute force contribution after ranking fluid points
by local VFM magnitude. At the displayed input frame ($t=0.073$~s) the top 10\%
contributes approximately 79\% of lift and 77\% of drag, and the top 20\%
approximately 90\% of both.}
  \label{fig:vfm_physics_basis}
\end{figure}

The VFM volume integral provides a second, geometrically distinct path to lift
and drag, supported on the wake and separation region rather than the body
wall. Let $\Omega$ be the fluid domain and $\mathbf e_k$ the wind-axis unit
vector for $k\in\{L,D\}$. VFM introduces a geometry- and wind-axis-dependent
auxiliary potential
\begin{equation}
  \begin{cases}
  \nabla^2\phi_k = 0, & \mathbf{x}\in\Omega,\\
  \nabla\phi_k\cdot \mathbf{n} = \mathbf{e}_k\cdot \mathbf{n},
    & \mathbf{x}\in\partial\Omega_b,\\
  \nabla\phi_k\rightarrow 0, & |\mathbf{x}|\rightarrow\infty ,
  \end{cases}
  \label{eq:vfm_aux_potential}
\end{equation}
from which the vortex-force factor is obtained as
\begin{equation}
  \boldsymbol{\Lambda}_k
  =
  \nabla^\perp\phi_k
  =
  (\phi_{k,y},-\phi_{k,x})^T
  \equiv
  (P_k,Q_k)^T .
  \label{eq:vfm_factor}
\end{equation}
VATO uses the vortex-pressure contribution of this formulation. Taking the
constant fluid density as $\rho=\rho_\infty$, the force per unit span is
\begin{equation}
  F_k^{\mathrm{VFM}}
  =
  \rho
  \int_{\Omega}
  \left(\boldsymbol{\Lambda}_k\cdot\mathbf{u}\right)\omega_z\,dA .
  \label{eq:vfm_force}
\end{equation}
With the freestream dynamic-pressure scale, the corresponding force coefficient
is
\begin{equation}
  C_k^{\mathrm{VFM}}
  =
  \frac{F_k^{\mathrm{VFM}}}{\frac{1}{2}\rho U_\infty^2 c}
  =
  \frac{1}{\frac{1}{2}U_\infty^2 c}
  \int_{\Omega}
  \left(\boldsymbol{\Lambda}_k\cdot\mathbf{u}\right)\omega_z\,dA ,
  \qquad k\in\{L,D\}.
  \label{eq:vfm_vp_coeff}
\end{equation}
The fields $P_L(x)$, $Q_L(x)$, $P_D(x)$, and $Q_D(x)$ are precomputed for each
geometry and angle of attack and sampled at the corresponding mesh points
(Figure~\ref{fig:vfm_physics_basis}a--d).

For physics validation and predicted-field post-processing,
Eq.~\eqref{eq:vfm_vp_coeff} is evaluated in discrete form on the query mesh.
The spanwise vorticity is reconstructed as
$w\equiv\omega_z=\partial v/\partial x-\partial u/\partial y$ using the native
mesh derivatives. Let $\mathcal{V}$ be the valid fluid-point set and $a_i$ the
nodal quadrature weight; in the sampled training path, $a_i$ also contains the
stratified-sampling correction. The discrete coefficient is
\begin{equation}
  C_{k}^{\mathrm{VFM}}
  =
  \frac{1}{\frac{1}{2}U_\infty^2 c}
  \sum_{i\in\mathcal{V}}
  \bigl[P_{k,i}u_i + Q_{k,i}v_i\bigr]\,
  \omega_i\,a_i,
  \qquad k\in\{L,D\}.
  \label{eq:Ck_vfm}
\end{equation}
For predicted-field readout, $(u_i,v_i,\omega_i)$ in
Eq.~\eqref{eq:Ck_vfm} are replaced by
$(\hat u_i,\hat v_i,\hat\omega_i)$, with $\hat\omega_i$ reconstructed from the
predicted velocity by the same mesh-curl operator.

Both operators are validated on the target fields before being applied to
predicted fields. The validation uses a representative test trajectory at an
incidence excluded from training, with
$(\theta_1,\theta_2)=(174.1^{\circ},155.0^{\circ})$ and AoA $=12^{\circ}$ over a
$0.18$~s physical time, and takes the $C_L$ and $C_D$ recorded by the solver as
the reference. The viscous contribution to these coefficients is negligible on
this case, so the reference is governed by the pressure force. The coefficients
recovered through Eqs.~\eqref{eq:F_press} agree closely with
the target values, with correlation coefficients $r=1.0000$ for $C_L$ and $r=0.9996$ for $C_D$. The VFM
integral, computed from $u$, $v$, and $\omega$ without any pressure information,
gives $r=0.9721$ and $r=0.9695$ and matches the target to mean absolute relative
errors of 2.09\% and 2.22\% (Figure~\ref{fig:vfm_physics_basis}e). Both
operators recover accurate coefficients from different information. The
agreement establishes both the accuracy of the VFM formulation on this flow and
the reliability of the two operators used to recover lift and drag from
predicted fields in Section~\ref{sec:results}.

The same integrand, evaluated pointwise on the target field by combining the
four panel-method coefficients with the reconstructed vorticity and quadrature
weight, gives the pointwise force-contribution maps $\Phi_{L,i}$ and
$\Phi_{D,i}$. At the displayed input-frame snapshot
($t=0.073$~s), the top 10\% of fluid points account for approximately 79\% of
the absolute lift contribution and 77\% of the absolute drag contribution;
the corresponding top-20\% shares are approximately 90\%
(Figure~\ref{fig:vfm_physics_basis}f). This concentration motivates the
fixed source budget in Section~\ref{subsec:goal-focus}.

\subsection{Data preprocessing}
\label{subsec:reynolds}

All CFD fields are first represented as point data on the native mesh. For each
geometry--incidence trajectory, an inlet-background state $\mathbf{b}_g$ is estimated by averaging
$[u,v,p]^T$ over the upstream band $\Omega_{\mathrm{up}}$ of the first frame.
The state field used for learning is the corresponding perturbation field,
\begin{equation}
  \tilde{\mathbf{y}}_i(t)
  =
  \mathbf{y}_i(t)-\mathbf{b}_g .
  \label{eq:state_residual}
\end{equation}
The residual state field and the spatial condition field are then standardised
using training-set statistics,
\begin{equation}
  \mathbf{y}_i^{*}(t)
  =
  \frac{\tilde{\mathbf{y}}_i(t)-\boldsymbol{\mu}_{y}}
  {\boldsymbol{\sigma}_{y}},
  \qquad
  c_i^{*}
  =
  \frac{c_i-\mu_{c}}
  {\sigma_{c}}.
  \label{eq:zscore_preprocess}
\end{equation}
Here $\tilde{\mathbf{y}}_i(t)$ is the perturbation state at point $i$,
$\mathbf{y}^{*}_i(t)$ and $c^{*}_i$ are its standardised counterpart and the
standardised condition field, and $\mu_y$, $\sigma_y$, $\mu_c$, $\sigma_c$ are
per-channel means and standard deviations computed over the training set.

The temporal scalars are scaled separately before being appended to the model
input. Points inside the aerofoil, identified from the signed-distance field, are
masked in the state and appended temporal channels, while signed distance remains
available as the geometry condition. Vorticity is not used as a predicted state
channel.

\subsection{VFM contribution field supervision}
\label{subsec:vfm-supervision}

VATO-S preserves the GAOT architecture and adds supervision on the two
per-point contribution fields whose sums give the discrete VFM $C_L$ and $C_D$. For $k\in\{L,D\}$, the VFM contribution field at mesh point $i$ is
\begin{equation}
  \Phi_{k,i}
  =
  \frac{(P_{k,i}u_i+Q_{k,i}v_i)\,\omega_i a_i}
       {\frac{1}{2}U_\infty^2c}.
  \label{eq:vfm-contribution}
\end{equation}
The auxiliary objective compares the predicted and target $C_L$/$C_D$ maps as
one vector-valued field,
\begin{equation}
  L^{\mathrm{VFM\mbox{-}contrib}}
  =
  \frac{
    \sum_{i\in\mathcal{V}}\sum_{k\in\{L,D\}}
    \left(\hat{\Phi}_{k,i}-\Phi_{k,i}\right)^2
  }{
    \max\!\left(
      \sum_{i\in\mathcal{V}}\sum_{k\in\{L,D\}}\Phi_{k,i}^2,
      \varepsilon\right)
  }.
  \label{eq:vfm-contribution-loss}
\end{equation}
The predicted contribution field $\hat\Phi$ uses predicted velocity and
curl-derived vorticity; pressure does not enter
Eq.~\eqref{eq:vfm-contribution}. The target field $\Phi$ is detached from the
gradient graph. Equation~\eqref{eq:vfm-contribution-loss} is evaluated for each
sample with $\varepsilon=10^{-8}$ and then averaged over the valid samples in
the batch. The target contribution field is available only during training. At
inference, VATO-S receives the same state,
geometry, and time inputs as the matched GAOT control and therefore adds no model parameters
or inference-time operator.

\subsection{Source prioritisation and residual cross attention}
\label{subsec:goal-focus}

VATO-A uses current flow vorticity together with VFM contribution and
sensitivity fields to identify regions relevant to Lift and Drag. This
information prioritises source locations but is neither embedded as an input
field nor used as an auxiliary loss. From the full native input frame, the
model combines the mesh curl of $(u,v)$, the precomputed VFM basis, nodal
quadrature weight,
and dynamic pressure to form signed local contributions and their normalised
positive and negative groups:
\begin{equation}
  \begin{aligned}
    \Phi_{k,i}
    &= \frac{(P_{k,i}u_i+Q_{k,i}v_i)\,\omega_i a_i}
             {\frac{1}{2}U_\infty^2c},
    && k\in\{L,D\},\\
    \sigma_k
    &= \left(\frac{1}{|\mathcal{V}|}
       \sum_{j\in\mathcal{V}}\Phi_{k,j}^{2}\right)^{1/2},
    &\widehat{\Phi}_{k,i}
    &= \operatorname{clip}\!\left(
       \frac{\Phi_{k,i}}{\max(\sigma_k,\varepsilon_r)},-\gamma,\gamma\right),\\
    r_{k,i}^{+}
    &= \max\!\left(\widehat{\Phi}_{k,i},0\right),
    &r_{k,i}^{-}
    &= \max\!\left(-\widehat{\Phi}_{k,i},0\right).
  \end{aligned}
  \label{eq:goal_focus_contribution}
\end{equation}
Here $\mathcal{V}$ is the valid fluid-point set, $\sigma_k$ is the framewise root mean square (RMS)
contribution, $\varepsilon_r=10^{-12}$ prevents division by zero, and
$\gamma=6$ is the fixed clipping limit. The four local VFM sensitivity factors,
defined with vorticity held fixed, are
\begin{equation}
  s_{k,u,i}
  = \frac{P_{k,i}\omega_i a_i}{\frac{1}{2}U_\infty^2c},
  \qquad
  s_{k,v,i}
  = \frac{Q_{k,i}\omega_i a_i}{\frac{1}{2}U_\infty^2c},
  \qquad k\in\{L,D\}.
  \label{eq:goal_focus_sensitivity}
\end{equation}
For $m\in\{u,v\}$, these factors are normalised and combined as
\begin{equation}
  \begin{aligned}
    \eta_k
    &= \left[\frac{1}{2|\mathcal{V}|}
       \sum_{j\in\mathcal{V}}
       \left(s_{k,u,j}^{2}+s_{k,v,j}^{2}\right)\right]^{1/2},\\
    \widehat{s}_{k,m,i}
    &= \operatorname{clip}\!\left(
       \frac{s_{k,m,i}}{\max(\eta_k,\varepsilon_r)},-\gamma,\gamma\right),\\
    r_i^{s}
    &= \left[\frac{1}{4}\sum_{k\in\{L,D\}}
       \sum_{m\in\{u,v\}}\widehat{s}_{k,m,i}^{2}\right]^{1/2},\\
    r_i^{\mathrm{all}}
    &= \left[(r_i^{s})^2+\widehat{\Phi}_{L,i}^{2}
       +\widehat{\Phi}_{D,i}^{2}\right]^{1/2}.
  \end{aligned}
  \label{eq:goal_focus_scores}
\end{equation}
The score $r_i^{s}$ forms the fifth prioritisation group, while
$r_i^{\mathrm{all}}$ fills any unassigned quota. All prioritisation values and
selected indices are detached, and only input-frame
quantities are used. Exactly 64 tokens are reserved for spatial coverage over
the valid fluid domain. The remaining 192-token budget is balanced over
$r_{L}^{+}$, $r_{L}^{-}$, $r_{D}^{+}$, $r_{D}^{-}$, and combined
VFM sensitivity,
with unfilled quotas assigned by the combined VFM contribution--sensitivity
score.

\begin{figure}[htbp]
  \centering
  \includegraphics[width=\linewidth]{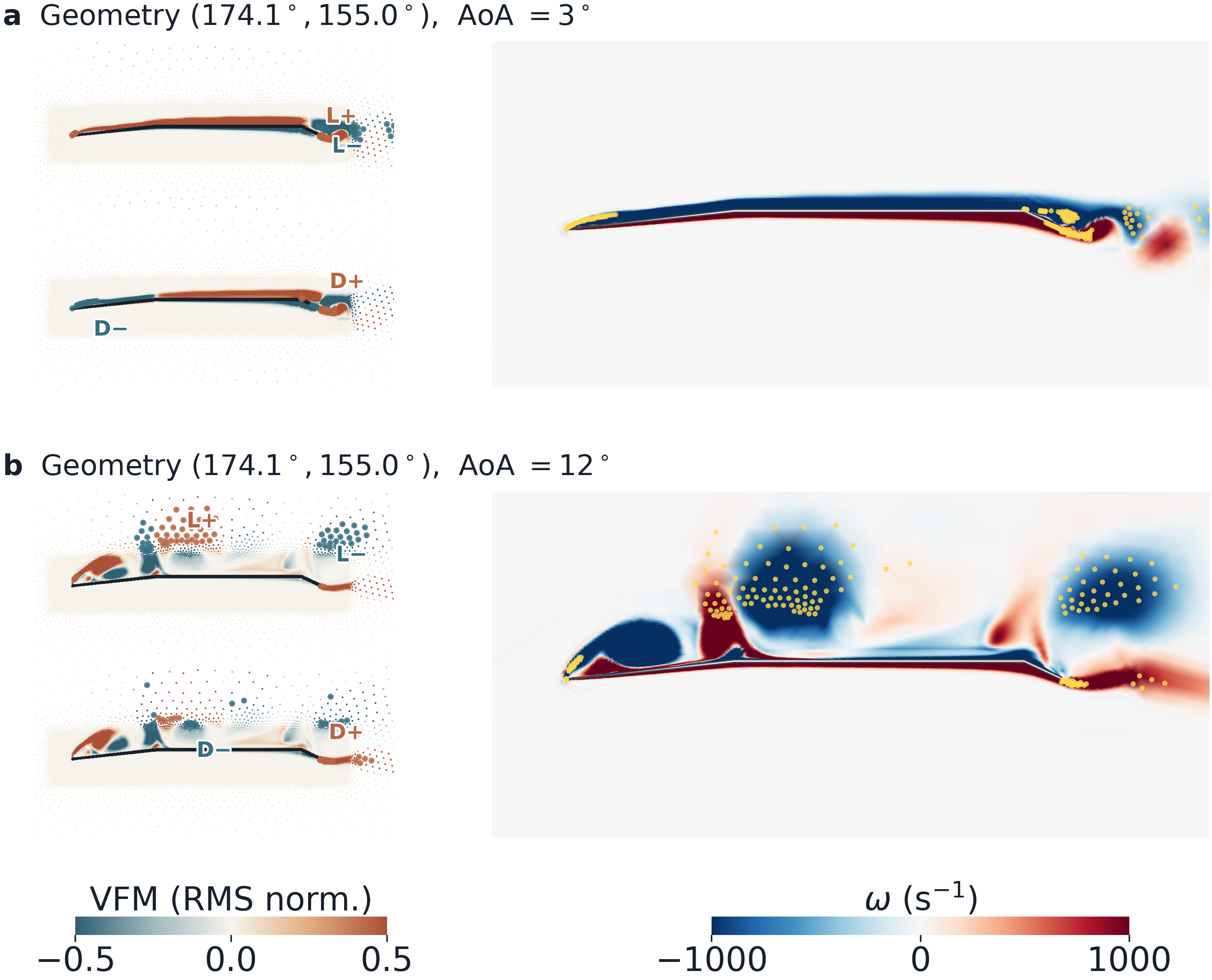}
  \caption{Source prioritisation for the input frame at $t=0.073$~s with
$(\theta_1,\theta_2)=(174.1^{\circ},155.0^{\circ})$, at (a)~AoA $=3^{\circ}$ and
(b)~$12^{\circ}$. The stacked left maps show the RMS-normalised signed VFM Lift
and Drag contributions, with markers identifying the indices retained from their
positive and negative groups; display values are clipped to $[-0.5,0.5]$ without
changing the indices. The right maps show the 192 VFM-prioritised indices over
the input-frame vorticity, with the 64 spatial-coverage indices omitted for
clarity. The displayed allocation uses the trained 256-token budget.
Prioritisation uses only detached input-frame quantities, and
the selected tokens carry ordinary state and coordinates rather than VFM values.
The vorticity background uses the common $[-1000,1000]$~s$^{-1}$ scale on the
native mesh.}
  \label{fig:goal-focus-allocation}
\end{figure}

Each selected token carries the baseline source features
$[u,v,p,c,t_{\mathrm{in}},\tau]$ and coordinates $(x,y)$, not the VFM values
used for prioritisation. Let $\mathbf{Z}_{F}$ denote these 256 full-frame tokens,
$\mathbf{S}_{\mathcal{N}(q)}$ the ordinary source tokens in the local
neighbourhood of query $q$, $\mathbf{H}$ the processed latent patches, $\mathbf{h}_q$ the MAGNO-decoded
feature at query $q$, and $\hat{\mathbf{y}}_q^{\,0}$ the preliminary UVP
prediction. VATO-A applies
\begin{equation}
  \begin{aligned}
    \mathbf{H}^{+}
    &= \mathbf{H}
       + \mathcal{A}_{\mathrm{patch}}(\mathbf{H},\mathbf{Z}_{F}),\\
    \mathbf{h}_{q}^{+}
    &= \mathbf{h}_{q}
       + \mathbf{g}_{q}\odot
         \mathcal{A}_{\mathrm{local}}
         (\mathbf{h}_{q},\mathbf{S}_{\mathcal{N}(q)}),\\
    \hat{\mathbf{y}}_{q}
    &= \hat{\mathbf{y}}_{q}^{\,0}
       + \mathcal{A}_{\mathrm{query}}
         ([\hat{\mathbf{y}}_{q}^{\,0},\mathbf{x}_{q}],\mathbf{Z}_{F}).
  \end{aligned}
  \label{eq:vato_main_attention_paths}
\end{equation}
Here each $\mathcal{A}$ is multi-head scaled dot-product attention, with its
first argument providing queries and its second providing keys and values;
$\odot$ denotes channel-wise multiplication. The operators
$\mathcal{A}_{\mathrm{patch}}$ and $\mathcal{A}_{\mathrm{query}}$ are eight-head,
width-192 residual cross attention modules. The former is applied after
Transformer processing and before MAGNO decoding; the latter is applied to
the preliminary UVP output in query chunks of 2{,}048. The local path uses
four-head, width-96 attention over the neighbourhood $\mathcal{N}(q)$ of at
most 32 ordinary source points within radius 0.04, including relative
displacement and distance. Its learned sigmoid gate
$\mathbf{g}_q\in(0,1)^{96}$ modulates the local residual before the final
decoder projection. All three
output projections are zero-initialised. Figure~\ref{fig:goal-focus-allocation}
visualises the prioritisation rule at the trained 256-token budget.

\subsection{Training objective and protocol}
\label{subsec:loss}

The GAOT reference, matched GAOT control, and both VATO variants optimise physical-space, squared relative
$L_2$ errors for velocity and pressure together with a mesh-curl vorticity
term:
\begin{equation}
  \begin{aligned}
    \Ltotal
    &= \sum_{f\in\{u,v,p\}}\alpha_f
      \frac{\|\hat f-f\|_{2,\mathcal{V}}^2}
           {\max(\|f\|_{2,\mathcal{V}}^2,\varepsilon_\ell)}
    + \alpha_w
      \frac{\|\hat w_{\log}-w_{\log}\|_{2,\mathcal{V}}^2}
           {\max(\|w_{\log}\|_{2,\mathcal{V}}^2,\varepsilon_\ell)},\\
    w_{\log}
    &= \operatorname{sign}(\omega)\log(1+|\omega|),\\
    \hat w_{\log}
    &= \operatorname{sign}(\hat\omega)\log(1+|\hat\omega|).
  \end{aligned}
  \label{eq:loss_total}
\end{equation}
where $\hat\omega$ is reconstructed from the mesh curl of the predicted
velocity and $\varepsilon_\ell=10^{-8}$. Each ratio is evaluated per sample over
valid fluid points, excluding body-interior and padded points, and then averaged
over the batch. Velocity and pressure are denormalised and the inlet background
is restored before this physical-space objective is evaluated. All four channel
weights are one. VATO-S uses
$L^{\mathrm{total}}+L^{\mathrm{VFM\mbox{-}contrib}}$ with unit weight on the
second term. The GAOT reference, matched GAOT control, and VATO-A use only
$L^{\mathrm{total}}$. Pressure-integral,
scalar force-coefficient, and VFM-integral losses have zero weight in all four
configurations and are used only as diagnostic readouts.

Optimisation uses AdamW with weight decay $1\times10^{-4}$, drop-path 0.1,
attention dropout 0.1, and a cosine-annealed learning rate from
$1\times10^{-4}$ to $5\times10^{-5}$. Training uses 8{,}192 sampled pairs per
epoch, batch size 8 per rank, and four-rank distributed data parallelism for
300 epochs. All reported comparisons use the final epoch-300 checkpoint. 

The four configurations are summarised in Table~\ref{tab:model-configurations}.
The GAOT reference uses uniform sampling, meaning the default training sampler
without trajectory reweighting. The matched GAOT control, VATO-S, and VATO-A
instead share flow-aware trajectory sampling. Each trajectory receives a fixed
flow-variation score $D_c$, formed from the spatial standard deviation of the
increment between source and target frames and weighted 0.20, 0.45, 0.25, and
0.10 over $u$, $v$, $p$, and the variation of that quantity between pairs, so
$D_c$ is governed by the transverse velocity and carries no vorticity
contribution. Source--target pairs are drawn with replacement in proportion to
$\max(0.05,\,D_c^{1/2})$, which every pair of a trajectory shares, so the number
of pairs a trajectory offers enters its sampling mass alongside $D_c$. The
scores are computed once before training, are not updated from model errors, and
are absent at inference. Because the matched control carries this sampler but no
vortex-force coupling, it isolates the effect of the coupling from that of the
sampler.

\begin{table}[htbp]
  \centering
  \caption{The four configurations. GAOT is the published architecture retrained
under the common task protocol, so the values listed come from this retraining
rather than from the cited paper. Uniform denotes the default training sampler
without trajectory reweighting; flow-aware denotes fixed training-only
trajectory weighting derived from the variation between source and target
frames. GAOT differs from GAOT (matched) only in the sampler, and GAOT
(matched) differs from VATO-S only in the contribution field loss, so each row
isolates one change from the row above. Inference is the median
preprocessing-plus-forward time per sample for batch size 8 on an NVIDIA H200,
excluding data loading and force post-processing.}
  \label{tab:model-configurations}
  \begingroup
  \fontsize{9.5pt}{11.5pt}\selectfont
  \setlength{\tabcolsep}{3.5pt}
  \renewcommand{\arraystretch}{1.16}
  \begin{tabularx}{\textwidth}{@{}
    >{\raggedright\arraybackslash}p{0.19\textwidth}
    >{\raggedright\arraybackslash}p{0.15\textwidth}
    >{\raggedright\arraybackslash}X
    >{\centering\arraybackslash}p{0.11\textwidth}
    >{\centering\arraybackslash}p{0.16\textwidth}@{}}
    \toprule
    Configuration & Training sampling & VFM coupling & Params. (M) & Inference (ms/sample) \\
    \midrule
    GAOT~\cite{ref23} & Uniform & -- & 19.09 & 34.9 \\
    GAOT (matched) & Flow-aware & -- & 19.09 & 34.8 \\
    \midrule
    VATO-S & Flow-aware & VFM contribution field supervision & 19.09 & 34.8 \\
    VATO-A & Flow-aware & VFM based source prioritisation and residual attention & 20.08 & 56.9 \\
    \bottomrule
  \end{tabularx}
  \endgroup
\end{table}

\subsection{Evaluation metrics}
\label{subsec:metrics}

Field errors are evaluated separately for the velocity vector,
pressure, and vorticity over the valid fluid-point set $\mathcal{V}$:
\begin{align}
  rL_{2}^{UV}
  &=
  \left[
  \frac{\sum_{i\in\mathcal{V}}
  \bigl((\hat u_i-u_i)^2+(\hat v_i-v_i)^2\bigr)}
  {\sum_{i\in\mathcal{V}}(u_i^2+v_i^2)}
  \right]^{1/2},
  \label{eq:rL2_uv}\\
  rL_{2}^{f}
  &=
  \left[
  \frac{\sum_{i\in\mathcal{V}}(\hat f_i-f_i)^2}
  {\sum_{i\in\mathcal{V}}f_i^2}
  \right]^{1/2},
  \qquad f\in\{p,w\}.
  \label{eq:rL2_scalar}
\end{align}
Both $w$ and $\hat w$ are reconstructed from the corresponding velocity field
with the same native mesh-curl operator. Tables and figures report
$100\,rL_2$ in percent. For operator
$o\in\{\mathrm{press},\mathrm{VFM}\}$ and force direction $k\in\{L,D\}$,
the sequence MAE for one source anchor and lead-time block $\mathcal{T}$ is
\begin{equation}
  \operatorname{MAE}_{k}^{o}
  = \frac{1}{|\mathcal{T}|}
    \sum_{\tau\in\mathcal{T}}
    \left|\hat C_{k}^{o}(\tau)-C_{k}^{o}(\tau)\right|.
  \label{eq:force_mae}
\end{equation}
Here $\hat C_k^o(\tau)$ and $C_k^o(\tau)$ apply the same operator $o$ to the
predicted and target fields, respectively, at lead time $\tau$.
For each reported lead-time window, pair-level field errors and anchor-level
force MAEs are averaged over anchors within a trajectory, over the six
incidences within a geometry, and then with equal weight over the nine
geometries. Lead-time-resolved figures retain $\tau$ before applying the same
anchor--incidence--geometry hierarchy. Body-interior points are excluded from
all field metrics.

\section{Results}
\label{sec:results}

\begin{figure}[H]
  \centering
  \includegraphics[width=\linewidth]{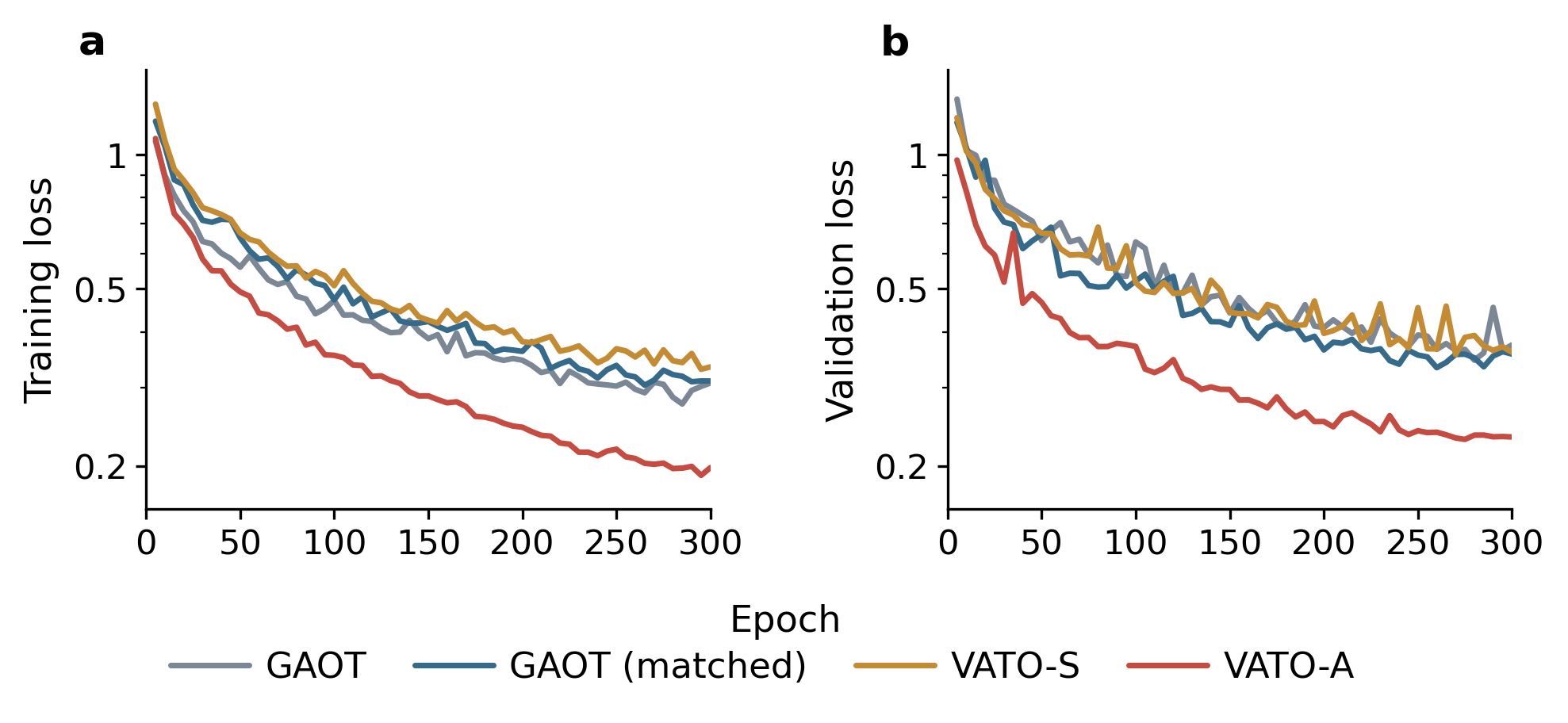}
  \caption{Training dynamics under the common 300-epoch protocol. (a)~Training and
(b)~validation values of the field objective shared by all four configurations,
defined by Eq.~\eqref{eq:loss_total} with unit channel weights. The auxiliary
contribution field term of VATO-S is excluded so that all four configurations
share the same ordinate. Each point is the exact aggregate logged at a
five-epoch interval, with no smoothing or interpolation.}
  \label{fig:training-dynamics}
\end{figure}

All four configurations reduce both the training and validation field objectives
under the common 300-epoch protocol (Figure~\ref{fig:training-dynamics}). The
GAOT reference ends at a validation field loss of 0.3741 and the matched GAOT
control at 0.3569. VATO-S ends at 0.3602, marginally above the matched control,
while the benchmark below shows it to be substantially more accurate over the
evaluated horizon; the validation view is restricted to a one-frame lead time,
so the contribution field target carries no measurable advantage on single-step
pointwise accuracy. VATO-A ends at 0.2325, a 34.9\% reduction relative to the
matched control. These curves describe one completed run per configuration.

\subsection{Comprehensive direct-field benchmark}
\label{subsec:generalisation}

The final 300-epoch checkpoints of the four configurations were applied to the
test set defined in Section~\ref{subsec:split}. Each of the 54 trajectories
supplies 28 source anchors and every integer lead time from 1 to 30~ms, giving
45{,}360 source--target combinations per configuration. Errors follow the
aggregation of Section~\ref{subsec:metrics} and are reported in
Table~\ref{tab:benchmark-field-accuracy}.

\begin{table}[htbp]
  \centering
  \caption{Field errors and force diagnostics on the 54-trajectory benchmark, with
45{,}360 source--target combinations per configuration. (a)~Fluid-only relative
$L_2$ error in percent, with velocity pooled over $u$ and $v$ through their
joint vector energy. (b)~Sequence mean absolute error in the $C_L$ and $C_D$
coefficients, obtained by applying the pressure-surface and vortex-force-volume
operators to the predicted and the target fields. Aggregation follows
Section~\ref{subsec:metrics}. Parentheses give the reduction against the GAOT
reference in the same column, with positive values favouring the named
configuration. Lead times 1--20~ms match the training horizon and 21--30~ms
extend it by 50\%. Values are shown to one decimal place, and column minima are
bold.}
  \label{tab:benchmark-field-accuracy}
  \newcommand{\BenchmarkReportedFieldRows}{%
    GAOT~\cite{ref24} & 15.44 & 18.83 & 31.42 & 18.82 & 24.10 & 31.83 \\
    GAOT (matched) & 15.56 {\scriptsize(-0.8\%)} & 20.60 {\scriptsize(-9.4\%)} & 31.65 {\scriptsize(-0.7\%)} & 19.31 {\scriptsize(-2.6\%)} & 25.82 {\scriptsize(-7.1\%)} & 32.47 {\scriptsize(-2.0\%)} \\
    VATO-S & 13.83 {\scriptsize(+10.4\%)} & 18.64 {\scriptsize(+1.0\%)} & 26.53 {\scriptsize(+15.6\%)} & 18.10 {\scriptsize(+3.8\%)} & 24.49 {\scriptsize(-1.6\%)} & 27.47 {\scriptsize(+13.7\%)} \\
    VATO-A & \textbf{12.99} {\scriptsize(+15.8\%)} & \textbf{17.41} {\scriptsize(+7.5\%)} & \textbf{21.62} {\scriptsize(+31.2\%)} & \textbf{17.60} {\scriptsize(+6.5\%)} & \textbf{24.09} {\scriptsize(+0.0\%)} & \textbf{23.27} {\scriptsize(+26.9\%)} \\
}
\newcommand{\BenchmarkReportedForceRows}{%
    \multirow{2}{*}{GAOT~\cite{ref24}} & Lift & 0.3791 & 0.3976 & 0.1725 & 0.2068 \\
     & Drag & 0.0455 & 0.0476 & 0.0205 & 0.0254 \\
    \midrule
    \multirow{2}{*}{GAOT (matched)} & Lift & 0.3693 {\scriptsize(+2.6\%)} & 0.3805 {\scriptsize(+4.3\%)} & 0.1388 {\scriptsize(+19.5\%)} & 0.1717 {\scriptsize(+17.0\%)} \\
     & Drag & 0.0436 {\scriptsize(+4.2\%)} & 0.0441 {\scriptsize(+7.3\%)} & 0.0165 {\scriptsize(+19.5\%)} & 0.0208 {\scriptsize(+18.0\%)} \\
    \midrule
    \multirow{2}{*}{VATO-S} & Lift & 0.3651 {\scriptsize(+3.7\%)} & 0.3762 {\scriptsize(+5.4\%)} & 0.1356 {\scriptsize(+21.4\%)} & 0.1477 {\scriptsize(+28.6\%)} \\
     & Drag & 0.0433 {\scriptsize(+4.7\%)} & 0.0447 {\scriptsize(+6.0\%)} & \textbf{0.0151} {\scriptsize(+26.7\%)} & \textbf{0.0169} {\scriptsize(+33.6\%)} \\
    \midrule
    \multirow{2}{*}{VATO-A} & Lift & \textbf{0.3262} {\scriptsize(+13.9\%)} & \textbf{0.3292} {\scriptsize(+17.2\%)} & \textbf{0.1241} {\scriptsize(+28.0\%)} & \textbf{0.1459} {\scriptsize(+29.4\%)} \\
     & Drag & \textbf{0.0377} {\scriptsize(+17.1\%)} & \textbf{0.0372} {\scriptsize(+21.7\%)} & 0.0164 {\scriptsize(+19.9\%)} & 0.0191 {\scriptsize(+24.9\%)} \\
}

  \setlength{\tabcolsep}{4.0pt}
  \renewcommand{\arraystretch}{1.12}
  \scriptsize
  \textbf{(a) Field prediction: relative $L_2$ error (\%)}\par\smallskip
  \resizebox{\textwidth}{!}{%
  \begin{tabular}{lcccccc}
    \toprule
    \multirow{2}{*}{Name}
          & \multicolumn{3}{c}{Lead time 1--20~ms}
          & \multicolumn{3}{c}{Lead time 21--30~ms} \\
    \cmidrule(lr){2-4}\cmidrule(lr){5-7}
          & Velocity & Pressure & Vorticity
          & Velocity & Pressure & Vorticity \\
    \midrule
    \BenchmarkReportedFieldRows
    \bottomrule
  \end{tabular}%
  }
  \par\medskip
  \begingroup
  \fontsize{10pt}{12pt}\selectfont
  \renewcommand{\scriptsize}{\fontsize{10pt}{12pt}\selectfont}
  \setlength{\tabcolsep}{1.8pt}
  \textbf{(b) $C_L$ and $C_D$ MAE from predicted fields}\par\smallskip
  \resizebox{\textwidth}{!}{%
  \begin{tabular}{lccccc}
    \toprule
    Name & Quantity
          & \shortstack{Pressure-derived\\1--20~ms}
          & \shortstack{Pressure-derived\\21--30~ms}
          & \shortstack{VFM-derived\\1--20~ms}
          & \shortstack{VFM-derived\\21--30~ms} \\
    \midrule
    \BenchmarkReportedForceRows
    \bottomrule
  \end{tabular}%
  }
  \endgroup
\end{table}

The two GAOT configurations separate the headline comparison from attribution
to the training protocol. The GAOT reference uses uniform sampling and is the denominator
for all displayed percentages. Relative to it, flow-aware sampling in the
matched GAOT control increases velocity, pressure, and vorticity error by
0.8\%, 9.4\%, and 0.7\% over lead times 1--20~ms and by 2.6\%,
7.1\%, and 2.0\% over lead times 21--30~ms. It nevertheless reduces every force MAE,
with gains ranging from 2.6\% to 19.5\%. Flow-aware sampling therefore lowers the force-diagnostic errors at the cost
of broad field accuracy. Both VATO
configurations retain this sampling regime.

Relative to the GAOT reference, VATO-S reduces velocity and vorticity error by
10.4\% and 15.6\% over lead times 1--20~ms and by 3.8\% and 13.7\% over lead
times 21--30~ms; its pressure changes are $+1.0\%$ and $-1.6\%$. Measured
against the matched GAOT control, which shares its sampling regime, the pressure
reductions are 9.5\% and 5.2\%, so the auxiliary term recovers the pressure
accuracy that flow-aware sampling gives up and, over lead times 1--20~ms,
returns it to the level of the reference. VATO-A gives reductions of
15.8\%, 7.5\%, and 31.2\% over lead times 1--20~ms and 6.5\%, 0.04\%, and 26.9\% over
lead times 21--30~ms. It is therefore the minimum-error configuration in all six field
cells, although its extrapolated-pressure advantage over the GAOT reference is
negligible. Paired geometry--incidence bootstrap intervals exclude zero for
five of the six VATO-A reductions; extrapolated pressure is the exception.

The force readouts separate the two configurations differently. VATO-A gives
the lowest pressure-derived $C_L$ and $C_D$ MAE in both windows, reducing them by
13.9\% and 17.1\% over lead times 1--20~ms and by 17.2\% and 21.7\% over lead times 21--30~ms.
It also reduces VFM-derived $C_L$ MAE by 28.0\% and 29.4\%. VATO-S instead
gives the lowest VFM-derived $C_D$ MAE, with reductions of 26.7\% and 33.6\%.
The two complete configurations therefore have distinct functional profiles:
VATO-A gives the stronger overall field and pressure-derived force result,
whereas VATO-S is most favourable for the VFM-derived Drag readout. Because
VATO-A changes source prioritisation, three residual attention paths, and
trainable capacity together, this contrast does not isolate the effect of the
VFM interface alone. 

The matched GAOT control and VATO-S each contain 19.09 million parameters and require
34.8~ms per sample. VATO-A contains 20.08 million parameters and requires
56.9~ms per sample, about 63.4\% above the matched control. The additional cost arises from its latent-patch, local-query, and output-query residual
attention paths.

\begin{figure}[htbp]
  \centering
  \includegraphics[width=\linewidth]{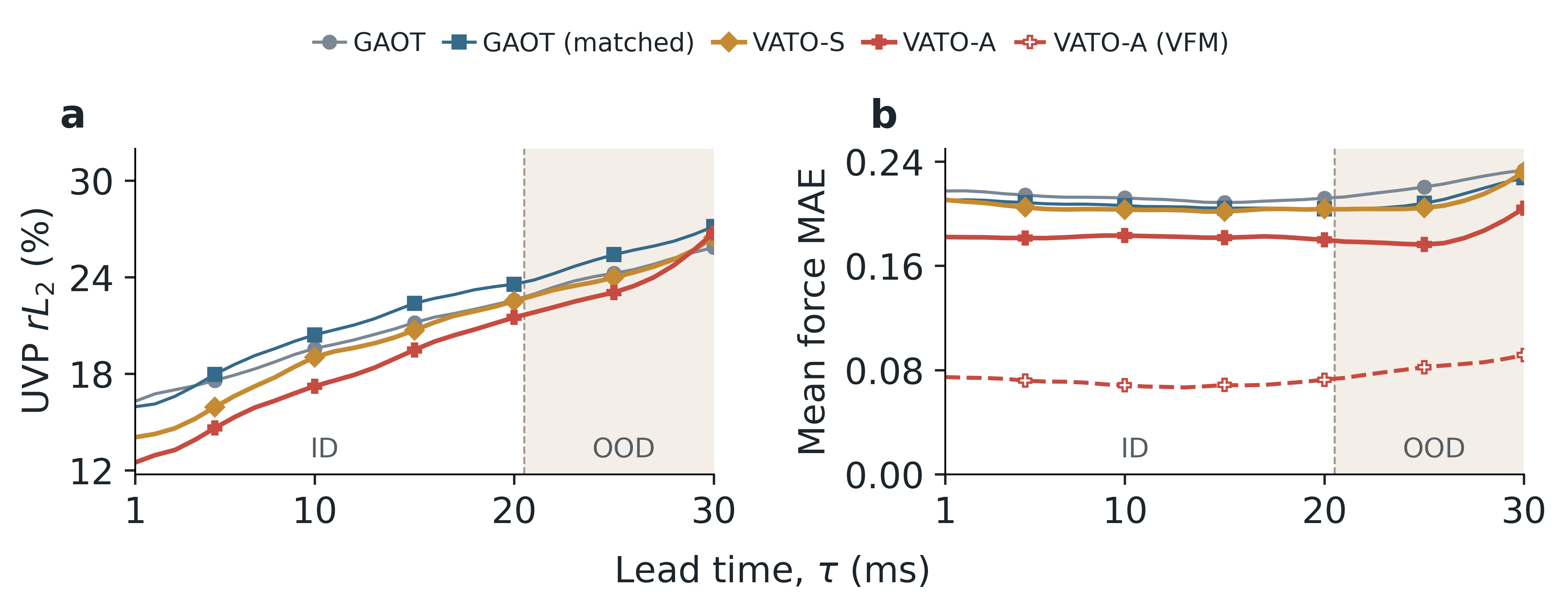}
  \caption{Lead-time-resolved absolute errors on the 54-trajectory population. At
each lead time, errors are averaged over the 28 anchors within each trajectory,
over the six evaluated incidences within each geometry, and then with equal
weight over the nine geometries. (a)~Equal-channel UVP relative $L_2$ error for
GAOT, GAOT (matched), VATO-S, and VATO-A. (b)~Equal mean of pressure-derived
$C_L$ and $C_D$ MAE at each lead time for all four configurations, with the
VFM-derived VATO-A result shown as an additional dashed curve. In-distribution (ID) training range ($\tau=1$--$20$~ms) and the temporal extrapolation beyond the maximum training lead-time, out-of-distribution (OOD) range ($\tau=21$--$30$~ms).}
  \label{fig:benchmark-lag}
\end{figure}

\begin{figure}[htbp]
  \centering
  \includegraphics[width=\linewidth]{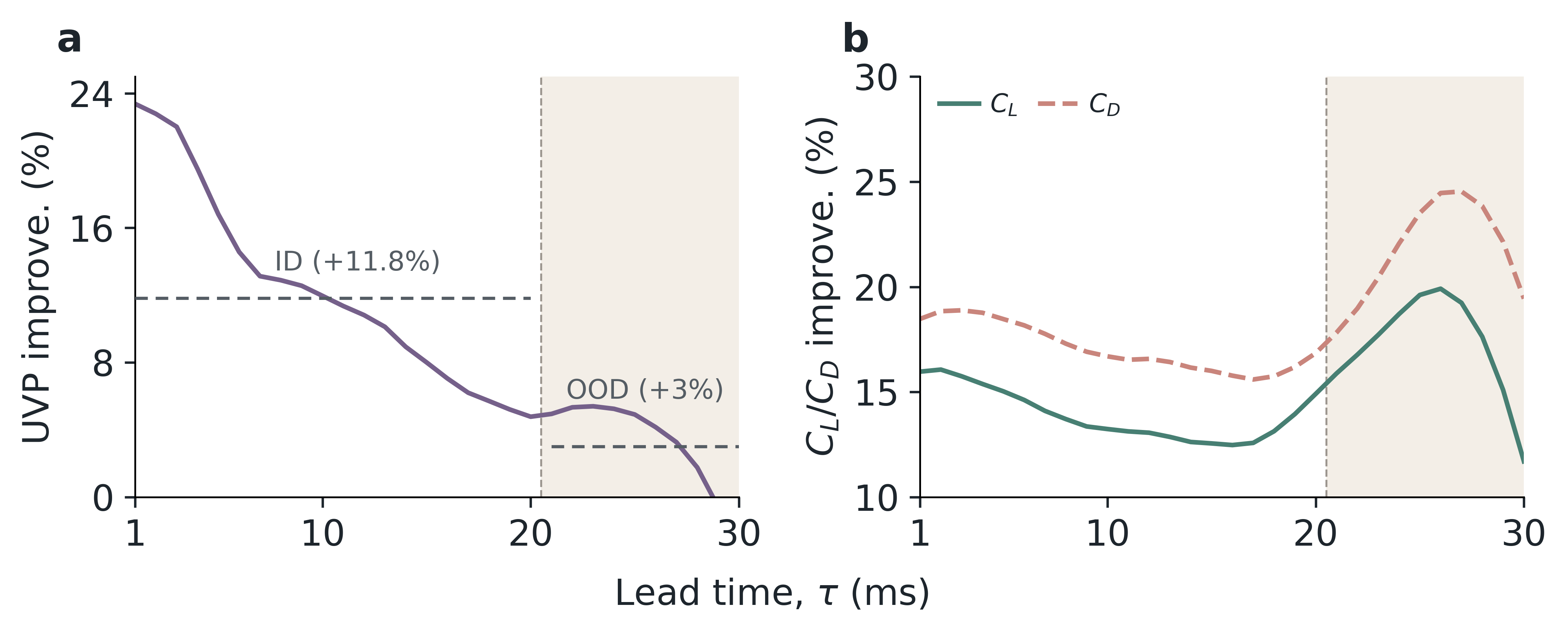}
  \caption{Lead-time-resolved improvement of VATO-A over the GAOT reference on
    the same reported 54-trajectory population. (a) Equal-channel UVP relative
    $L_2$ improvement. (b) Pressure-derived $C_L$ and $C_D$ MAE improvement.
    Improvement is defined as
    $100(E_{\mathrm{GAOT}}-E_{\mathrm{VATO\text{-}A}})
    /E_{\mathrm{GAOT}}$, where $E$ denotes UVP relative $L_2$
    error in (a) and the corresponding $C_L$ or $C_D$ MAE in (b); positive values
    favour VATO-A. The dashed horizontal segments in (a) report window-level
    improvement after averaging each model's error over the ID and OOD.}
  \label{fig:benchmark-lag-improvement}
\end{figure}

Resolving the same errors by lead time shows the ordering of the four
configurations to be stable across the horizon rather than produced by a
particular window (Figure~\ref{fig:benchmark-lag}).
Figure~\ref{fig:benchmark-lag-improvement} resolves the improvement of VATO-A
over the reference by lead time. Panel~(a) reports the equal-channel UVP
relative $L_2$ improvement, which reaches close to 24\% at the shortest lead
times and decreases as the horizon grows, giving window-level values of 11.8\%
over the trained range and 3.0\% over the extrapolation window. Panel~(b)
reports the improvement in the pressure-derived $C_L$ and $C_D$ mean absolute
error on the same population, which holds between roughly 12\% and 20\% across
the trained range.

Beyond 20~ms the two force curves rise together rather than following the field
metric downward, reaching close to 20\% for $C_L$ and close to 25\% for $C_D$ near
25~ms before falling back toward 30~ms. $C_L$ and $C_D$ are
generated by the shed vortices, so these readouts respond to how well the
coherent structures are preserved rather than to the pointwise accuracy of the
field as a whole. The rise indicates that the reference reproduces those
structures markedly less well once the requested lead time leaves the range seen
in training, while VATO-A sustains them over part of the extrapolation window,
and the subsequent fall toward 30~ms follows as the prediction of both
configurations degrades further.

\subsection{Aerodynamic functional consistency}
\label{subsec:benchmark-force}

\begin{figure}[htbp]
  \centering
  \includegraphics[width=\linewidth]{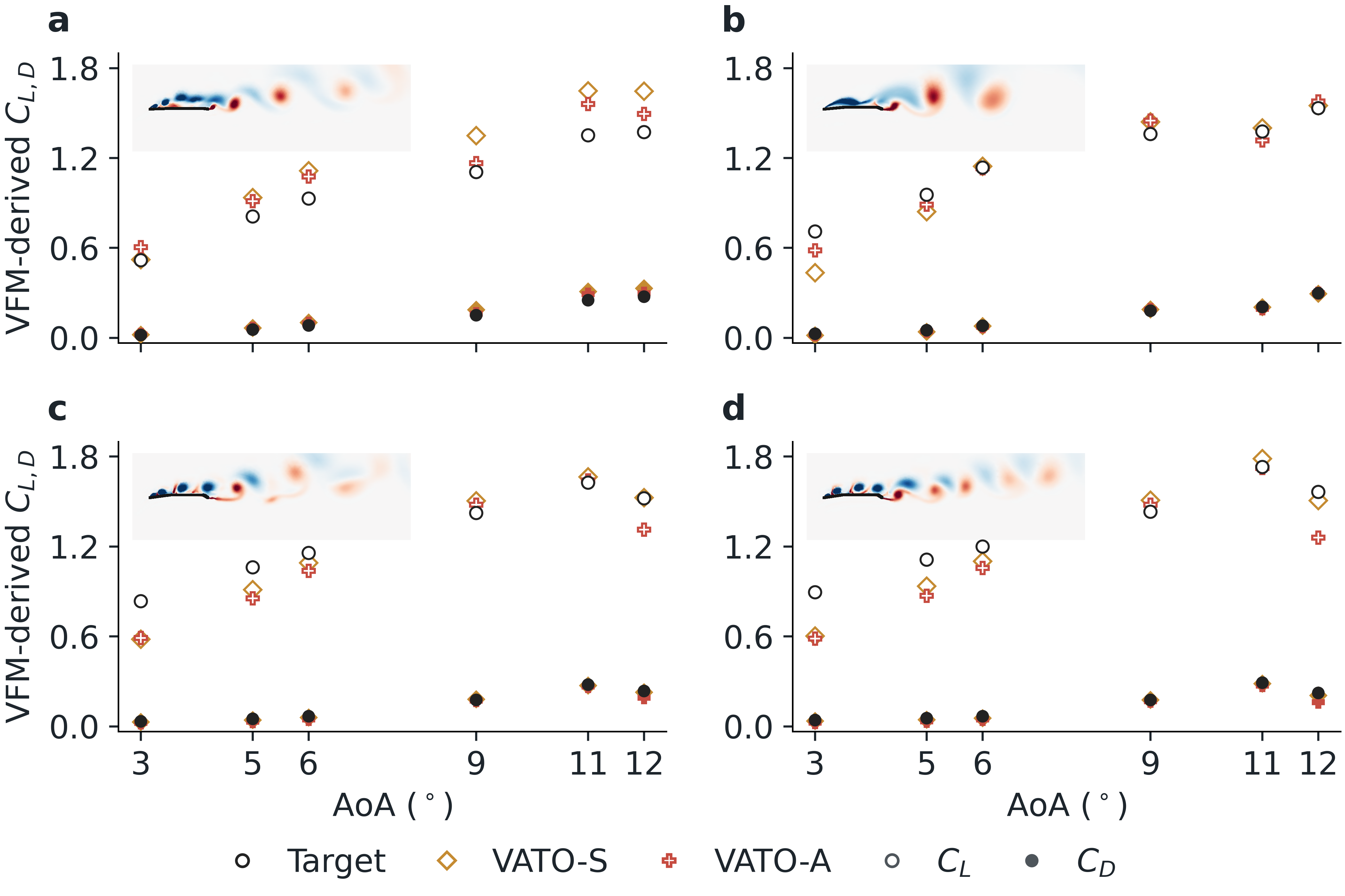}
  \caption{Fixed-geometry VATO predictions of VFM-derived mean $C_L$ and $C_D$
against matched targets. (a--d)~The thinnest, median, second-thickest, and
thickest geometries of the nine geometry sweep, with
$(\theta_1,\theta_2)$ equal to $(178.32^\circ,172.41^\circ)$,
$(174.95^\circ,158.19^\circ)$, $(172.46^\circ,149.04^\circ)$, and
$(171.63^\circ,146.31^\circ)$. Within each geometry and AoA, each point is the
mean force coefficient over all direct predictions at lead times 1--30~ms.
Circles are the target values; diamonds and crosses are the mean force coefficients predicted by VATO-S and VATO-A. Open markers denote $C_L$ and filled markers denote $C_D$.
The overlay in each panel shows a target-vorticity snapshot at AoA
$=12^\circ$ for the same geometry, rendered on the native mesh with the common
$w\in[-1000,1000]$~s$^{-1}$ range.}
  \label{fig:benchmark-force-polars}
\end{figure}

The force MAEs of Table~\ref{tab:benchmark-field-accuracy} pool over geometry
and incidence. Figure~\ref{fig:benchmark-force-polars} instead resolves the
VFM-derived $C_L$ and $C_D$ by incidence for four geometries spanning
the thickness range of the reported population, with each point a
benchmark-window mean over the same 28 source anchors and direct lead times
1--30~ms.

The four fixed-geometry views expose incidence-dependent behaviour that an
across-geometry coefficient average would hide. Both VATO variants reproduce
the overall $C_L$ rise and the lower-amplitude $C_D$ trend, including the
high-incidence turning of the thicker geometries. Agreement remains
geometry-dependent: VATO-A is generally closer for the thinnest geometry,
whereas VATO-S more closely follows the AoA $=12^\circ$ response of the two
thickest geometries. 

\subsection{Representative and diagnostic field predictions}
\label{subsec:benchmark-fields}

\begin{figure}[p]
  \centering
  \includegraphics[width=\linewidth]{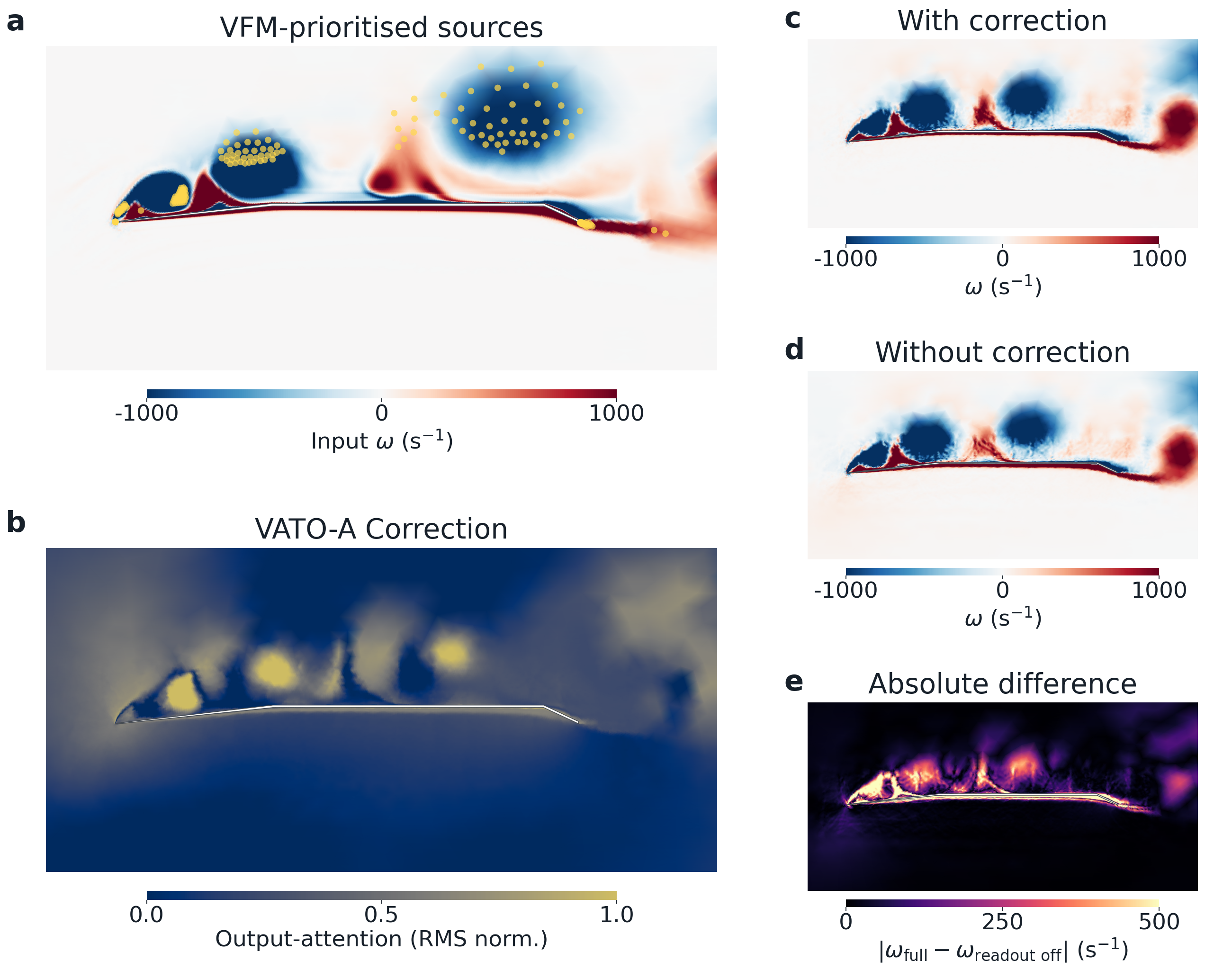}
  \caption{Output-level residual cross-attention correction in VATO-A for one
input from the test population, with the geometry, AoA, source time, and lead
time fixed in advance. The geometry is
$(\theta_1,\theta_2)=(174.1^{\circ},155.0^{\circ})$ at AoA $=12^{\circ}$, with
source time 0.081~s and lead time $\tau=10$~ms. (a)~Current-input vorticity $w$
with the realised VFM-prioritised source locations at the trained budget of 256
tokens; the spatial-coverage locations are omitted from the display.
(b)~RMS magnitude of the output-level residual correction across the normalised
$[u,v,p]$ channels, linearly stretched between its fluid-point 5th and 99th
percentiles. (c)~VATO-A prediction with the output correction. (d)~Prediction
from the same checkpoint after bypassing only the selected-token global
query-readout residual, with the latent residual and the local geometric
attention left active. (e)~Pointwise absolute vorticity difference between (c)
and (d). Vorticity panels use $[-1000,1000]$~s$^{-1}$ and the difference panel
$[0,500]$~s$^{-1}$; clipping affects the display only.}
  \label{fig:vato-a-correction-diagnostic}
\end{figure}

Figure~\ref{fig:vato-a-correction-diagnostic} examines the output-level residual
path on a single frame with the geometry, AoA, source time, and lead time fixed
in advance. Panel~(a) shows the prioritisation rule of Figure~\ref{fig:goal-focus-allocation}
realised at the trained budget of 256 tokens, with the selected locations lying
on the vortices above the section and in the near wake. In panel~(b), the RMS
magnitude of the output correction is distributed along the upper surface and
coincides with the vortices convected downstream from the sharp leading edge,
which generate the suction that carries the load. Bypassing this path on the
same checkpoint, with the latent residual and the local attention left active,
raises the vorticity relative $L_2$ error from 58.4\% to 64.2\%, and the
resulting difference in panel~(e) attains its maximum along the same surface,
averaging 368.1~s$^{-1}$ within the displayed crop. These values apply to this
intervention on this frame.

\begin{figure}[H]
  \centering
  \includegraphics[width=\linewidth,height=0.73\textheight,keepaspectratio]{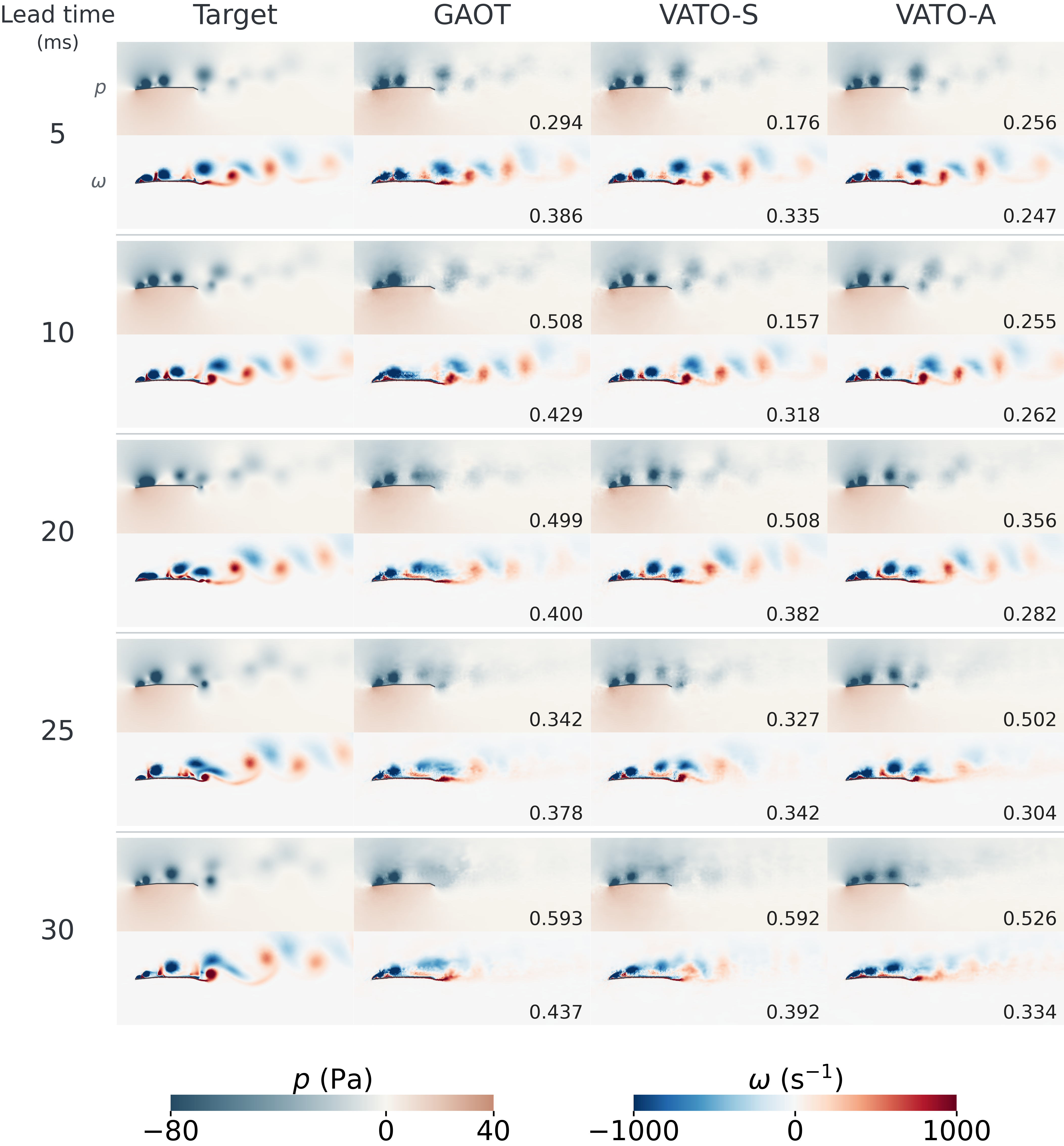}
  \caption{Predictions of the pressure and vorticity fields by the three models
for the same case, $(\theta_1,\theta_2)=(174.1^{\circ},155.0^{\circ})$ at AoA
$=12^{\circ}$, with a source time of $0.081$~s. Rows show lead times of 5, 10,
20, 25, and 30~ms; columns compare the target with the three predictions. Each
cell places the pressure field above the vorticity field, and the numbers on the
prediction tiles give the unclipped relative $L_2$ error over the valid fluid
points. All panels use $x/c\in[-0.30,3.50]$ and $y/c\in[-0.73,0.77]$, with
colour ranges $p\in[-80,40]$~Pa and $w\in[-1000,1000]$~s$^{-1}$.}
  \label{fig:fixed-geometry-leadtime-fields}
\end{figure}

The three models are compared at $12^{\circ}$, an incidence held out from both
training and validation, at fixed geometry
(Figure~\ref{fig:fixed-geometry-leadtime-fields}). The frame at $t=0.081$~s
serves as the source, and predictions are made 5, 10, 20, 25, and 30~ms ahead of
it, with the pressure field in the upper row and the vorticity field in the
lower row of each block. The baseline shown alongside the two VATO
configurations is GAOT rather than the matched control, GAOT being the more
accurate of the two on every field metric of
Table~\ref{tab:benchmark-field-accuracy}. All three models predict the shed
structures at a lead time of 5~ms. From a lead time of 10~ms onwards the GAOT
prediction degrades visibly: adjacent cores on the upper surface are no longer
resolved as separate structures and merge, so the predicted vorticity
distribution departs from the target in shape while its overall position is
retained. The vortex shed downstream of the trailing edge is still predicted by
GAOT at 20~ms, although with a markedly different shape, and is absent at
25~ms, whereas both VATO configurations continue to resolve it at 25 and 30~ms.
The vorticity errors follow the same ordering at every displayed lead time:
VATO-S is below GAOT throughout, and VATO-A is the most accurate of the three,
with an error of 0.334 at 30~ms against 0.392 for VATO-S and 0.437 for GAOT. In
pressure the ordering is less uniform, VATO-A being the most accurate at 20 and
30~ms and VATO-S at the three remaining lead times. 

\begin{figure}[p]
  \centering
  \includegraphics[width=\linewidth,height=0.80\textheight,keepaspectratio]{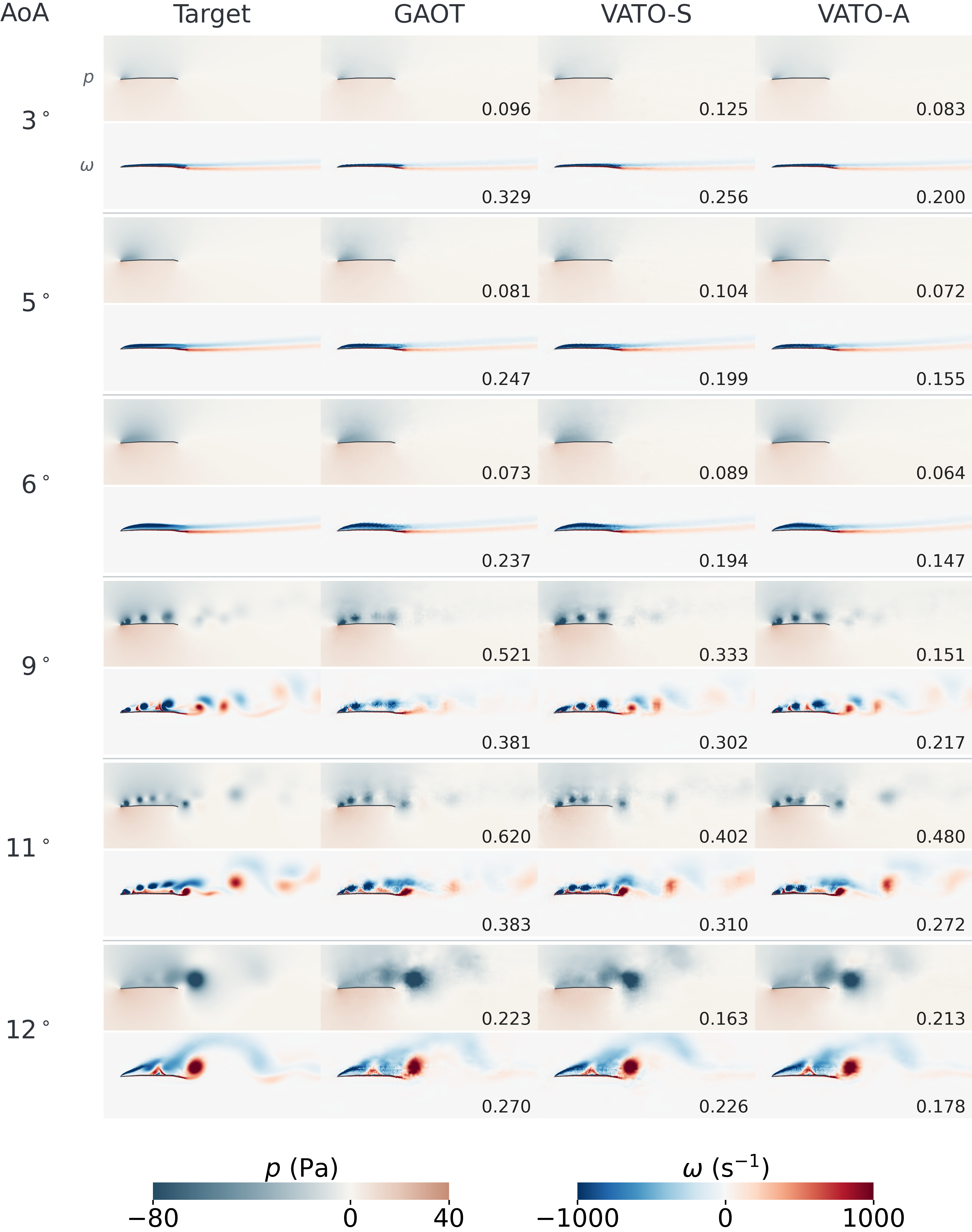}
  \caption{Predictions by the three models for the same geometry,
$(\theta_1,\theta_2)=(176.6^{\circ},165.1^{\circ})$, at a lead time of 20~ms.
Rows span the six test incidences; columns compare the target with the three
predictions. Each cell places the pressure field above the vorticity field, and
the numbers on the prediction tiles give the unclipped relative $L_2$ error over
the valid fluid points. All panels share $x/c\in[-0.30,3.50]$ and
$y/c\in[-0.73,0.77]$, with colour ranges $p\in[-80,40]$~Pa and
$\omega\in[-1000,1000]$~s$^{-1}$.}
  \label{fig:fixed-geometry-aoa-sweep}
\end{figure}

\begin{figure}[htbp]
  \centering
  \includegraphics[width=\linewidth]{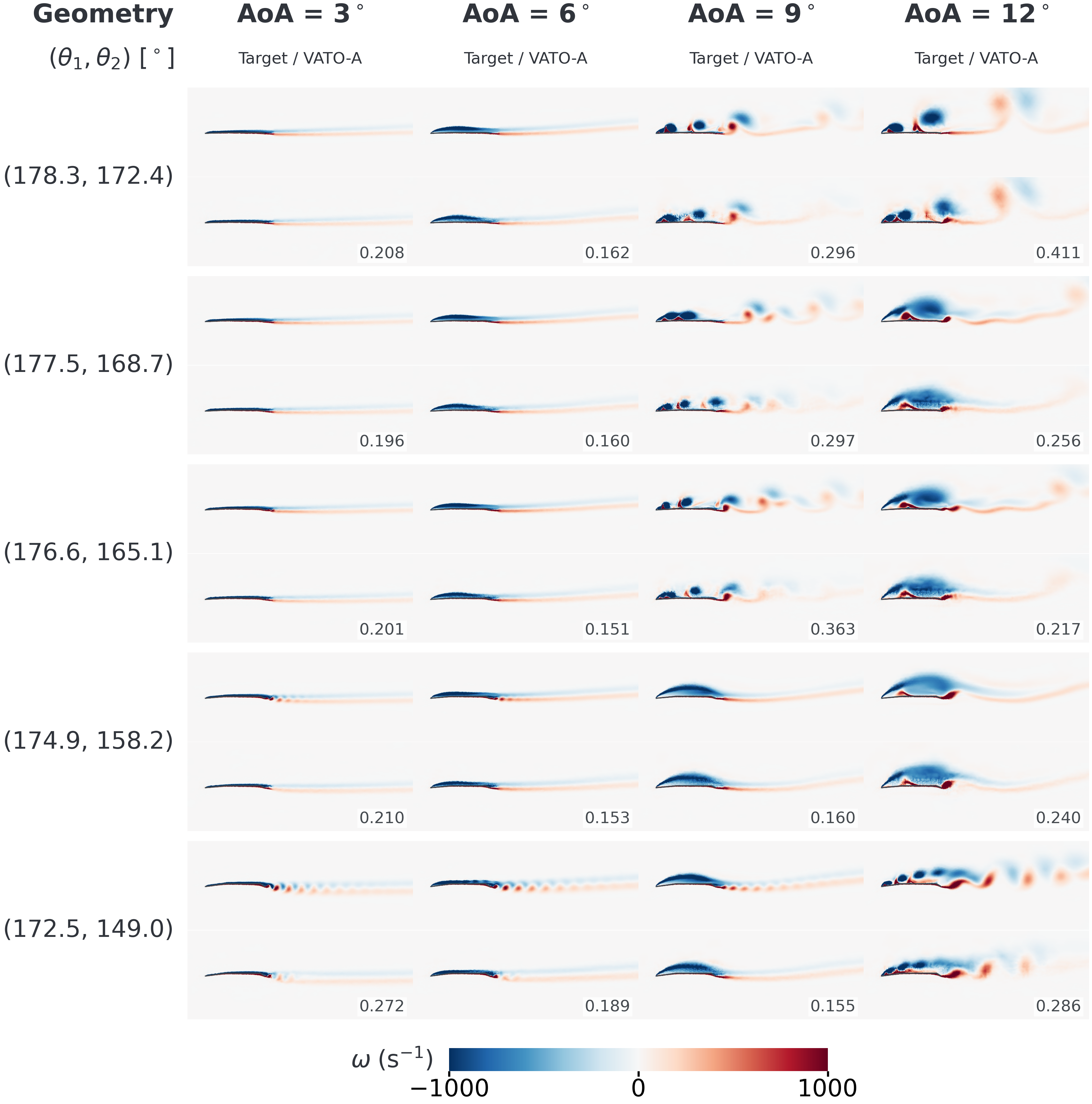}
  \caption{VATO-A predictions of the vorticity field across geometry and
incidence at a lead time of 20~ms. Rows show five randomly selected test geometries
indexed by the interior fold angles $(\theta_1,\theta_2)$; columns show four incidences. Each cell places
the target vorticity above the VATO-A prediction, and the number on each
prediction gives the unclipped relative $L_2$ error over the valid fluid points.
All panels share $x/c\in[-0.30,3.50]$, $y/c\in[-0.73,0.77]$, and
$w\in[-1000,1000]$~s$^{-1}$.}
  \label{fig:geometry-aoa-vorticity-atlas}
\end{figure}

Figure~\ref{fig:fixed-geometry-aoa-sweep} shows the same comparison across the
six test incidences at a fixed lead time of 20~ms. As the flow develops from
attached to separated and the prediction becomes more demanding, VATO-A is more
accurate than GAOT in both fields at every incidence, by 29\% to 43\% in
vorticity and by up to 71\% in pressure at $9^{\circ}$. VATO-S also resolves the
vortex structures that GAOT fails to recover, improving on it in vorticity at
every incidence and in pressure once the flow separates. VATO-A, the most
accurate of the three on the vorticity field, is examined against the target
over five randomly selected geometries and four incidences
(Figure~\ref{fig:geometry-aoa-vorticity-atlas}), which extends the comparison
from single cases to a broader sample of the population. The twenty cells cover
the flow states the family spans: a thin attached shear layer along the surface,
a regular vortex street in the wake downstream of the trailing edge, a single
large separated region over the upper surface, and discrete cores arranged along
the upper surface. VATO-A predicts each of them, together with the transitions
between them that occur as incidence increases within a geometry and as the fold
angles change at fixed incidence. The relative $L_2$ errors range from 0.151 to
0.411, the largest values arising where several discrete cores must be
positioned individually.

\clearpage

\section{Discussion and Limitations}
\label{sec:discussion}

The training sampler and the vortex-force coupling act on different quantities.
Relative to the uniformly sampled reference, the matched control raises every
field error and lowers every force MAE, so flow-aware sampling changes the
balance between the two families of metric rather than giving a uniformly
stronger configuration. Both VATO configurations improve on that control in
velocity, pressure, and vorticity across both lead-time windows, which
attributes the field improvement to the coupling rather than to the sampler.

Each interface improves the functional it is aligned with. VATO-S is trained on
the contribution field and gives the lowest VFM-derived Drag error, at the
parameter count and inference cost of the backbone. VATO-A is trained on neither
functional and improves the predicted field broadly, including pressure, so both
readouts follow and it gives the lowest pressure-derived $C_L$ and $C_D$ errors,
at about 63\% more measured inference time. The choice between them follows from
which functional is required and how much inference time is available: VATO-S
adds nothing at inference, while VATO-A spends additional communication and
capacity on the predicted field.

Outside the trained lead-time range the two families of metric separate again.
Over the 50\% temporal extension, the pointwise margins of VATO-A fall, its
vorticity margin is retained at 26.9\%, and all four force margins increase.
Lift and drag are generated by the shed vortices, so the quantities that hold
their advantage there are those that depend on the coherent structures rather
than on the pointwise accuracy of the field as a whole. A surrogate placed in a
design or control loop is queried at horizons its training set did not fix, and
the readouts it is asked to supply are the force functionals.

Where each interface enters bounds what can be inferred from it. The
contribution field constrains force-relevant combinations of velocity and
vorticity, does not uniquely determine the underlying state, and contains no
pressure term, so its improvements do not establish that supervising it alone
recovers every force-bearing flow feature. In VATO-A the VFM values determine
only which locations are prioritised and are not embedded in learned query, key,
or value content, but the reported comparison changes the prioritisation rule,
the residual attention paths, and the trainable capacity together. Its result is
an effect of the complete configuration, and isolating the rule would require a
geometry-only prioritisation control matched in path and capacity.

The reported population contains 54 trajectories from nine geometries
represented during training at other incidences, so the evidence covers unseen
incidences on known geometries rather than unseen shapes. Thirty-six of these
trajectories also provided the 1-ms validation view, and all comparisons use
frozen final checkpoints, so no selection was performed on them. Every
configuration is represented by one training seed, and the hierarchical
bootstrap quantifies variation over geometry and incidence rather than over
retraining. Training encodes 12{,}000 sampled source points whereas the
benchmark supplies the full native mesh to every configuration. The force
results are matched-operator diagnostics, with the same operator applied to
predicted and target fields, and both operators are inviscid in construction and
recover the pressure-derived component of the load. Multiple seeds, unseen
geometries, cross-solver tests, calibrated force validation, and autoregressive
evaluation would be needed to support stronger claims.

\section{Conclusion}
\label{sec:conclusion}

We introduced VATO, a family of neural operators that couples the Vortex Force
Map method to a geometry-aware transformer backbone at two interfaces. The
coupling draws on a geometry-only auxiliary potential problem and requires no
additional flow solution. It was evaluated on double-edged-plate aerofoils at
incidences excluded from training, against a retrained backbone reference and a
sampling-matched control.

VATO-S supervises the per-point contribution field during training, leaving the
architecture, the parameter count, and the inference cost of the backbone
unchanged. Relative to the reference it reduces velocity and vorticity error by
10.4\% and 15.6\% over the trained lead-time range, and it gives the lowest
VFM-derived Drag error of the family. VATO-A uses the same information to
prioritise 256 source locations for residual cross attention, reduces velocity,
pressure, and vorticity error by 15.8\%, 7.5\%, and 31.2\% over the same range,
and gives the lowest pressure-derived $C_L$ and $C_D$ errors, at about 63\% more
measured inference time.

Over a horizon 50\% longer than the longest lag seen in training, the pointwise
margins of VATO-A fall while its vorticity margin is retained at 26.9\% and all
four force margins increase. The advantage outside the trained range therefore
lies in the quantities that carry the load. The sampling-matched control
separates a second effect along the same line, in which flow-aware sampling on
its own lowers force error and raises pointwise field error.

The two interfaces cover a range of deployment conditions, from a training-only
intervention that leaves inference unchanged to one that spends additional
inference time on the predicted field. Separating the contribution of the
prioritisation rule from that of the additional attention paths, and testing on
unseen geometries, are the next steps.

\section*{Acknowledgements}
This research was funded by the Engineering Start-up Grant from King's College
London and by the Daiwa Anglo-Japanese Foundation through Daiwa Foundation
Awards (14465/15310).  The authors acknowledge the use of King's Computational
Research, Engineering and Technology Environment (CREATE) in conducting this
research. The authors appreciate financial support from the China Scholarship Council Program (202508440025)   

\section*{Declaration of generative AI and AI-assisted technologies in the
writing process}

During the preparation of this work the authors used generative AI tools to
assist with language editing and manuscript structuring. The authors reviewed
and edited the content as needed and take full responsibility for the content
of the publication.

\end{document}